\documentclass[11pt]{article}

\usepackage[preprint]{acl}

\usepackage{times}
\usepackage{latexsym}
\usepackage{multirow}
\usepackage{amsmath}
\usepackage{amssymb} 

\usepackage[T1]{fontenc}
\usepackage[utf8]{inputenc}

\usepackage{microtype}

\usepackage{inconsolata}

\usepackage{graphicx}

\usepackage{tcolorbox}
\tcbuselibrary{breakable, skins, listings}
\usepackage{listings}
\usepackage{makecell}

\newtcblisting{promptbox}{
  colback=gray!5,
  colframe=black!30,
  boxrule=0.5pt,
  arc=2pt,
  left=4pt,
  right=4pt,
  top=4pt,
  bottom=4pt,
  width=\columnwidth,
  breakable,
  enhanced,
  listing only,
  listing options={
    basicstyle=\ttfamily\footnotesize,
    breaklines=true,        % 自动换行
    breakatwhitespace=false
  }
}

\usepackage[table]{xcolor}
\definecolor{grey_purple}{RGB}{225,225,245}
\definecolor{highlight}{RGB}{255, 235, 210}
\definecolor{grey_blue}{RGB}{235,240,248}
\definecolor{grey}{RGB}{240,240,240}
\usepackage{booktabs}

\title{GIFT: Games as Informal Training for Generalizable LLMs}

\author{
  \textbf{
  Nuoyan Lyu$^{1,2}$, 
  Bingbing Xu$^{1 }$\thanks{Corresponding author.},
  Xueyun Tian$^{1,2}$,
  Weihao Meng,
  Yige Yuan$^{3}$, 
  } \\
  \textbf{Yang Zhang$^{4}$,Zhiyong Huang$^{4}$, Tat-Seng Chua$^{4}$, 
  Huawei Shen$^{1,2}$}
  \\
  $^{1}$ State Key Laboratory of AI Safety, Institute of Computing Technology, CAS \\
  $^{2}$ University of Chinese Academy of Sciences \\
  $^{3}$ University of Washington \\
  $^{4}$ National University of Singapore \\
  \texttt{\{lvnuoyan23z,xubingbing,tianxueyun23z\}@ict.ac.cn, wmeng9@jh.edu} 
}

\begin{document}
\maketitle
\begin{abstract}

Recent LLMs excel at formal tasks such as mathematical reasoning and code generation, but still struggle with broader abilities such as planning, creativity, and social intelligence. 
Inspired by human learning, where formal instruction and informal experience jointly shape intelligence, we introduce informal learning into LLM training and use games as annotation-free, feedback-driven environments. To cover diverse abilities including abstract reasoning, planning, creativity, and social interaction, we combine formal math tasks with three representative game tasks, including Matrix Games, TicTacToe, and Who's the Spy.
However, directly mixing these tasks under a unified RL objective can blur task-specific learning signals and provides no explicit guidance for coordinating task-gradient directions. 
To combat these, we propose Coordinated Subtask Training (CST), which replaces a single mixed update with sequential subtask-specific updates, separating heterogeneous RL signals while implicitly promoting coordination among subtasks. 
Experiments on ability-oriented benchmarks show that game-based informal learning improves generalization beyond formal training alone, while CST further enhances multi-task RL by preserving in-domain subtask performance and improving broader general abilities. Code and data are publicly available\footnote{\url{https://anonymous.4open.science/r/GIFT-LLM}}.

\end{abstract}

\section{Introduction}

Human intelligence arises from the interplay between formal and informal learning \cite{scribner1973cognitive}. Formal learning targets structured, task-specific knowledge, whereas informal learning builds practical wisdom through interaction, experience, and feedback \cite{callanan2011informal}. % Formal learning is structured and goal-oriented, typically targeting task-specific knowledge. In contrast, informal learning arises from unstructured interactions, iterative experience, and implicit feedback, fostering practical wisdom and transferable intelligence \cite{callanan2011informal}. %Formal learning emphasizes structured, goal-oriented education for acquiring task-specific knowledge, whereas informal learning unfolds in everyday environments through the unstructured interactions involving iterative experience and implicit feedback, %the unstructured interactions with iterations of practice and feedback, 
%txy 这两段是不是可以再合并一下？
In parallel, recent large language models (LLMs) have achieved remarkable success on formal learning tasks \cite{liu2025reinforcement, xu2025towards}, including mathematical reasoning \cite{wang2025reinforcement, zhang2025deeptheorem, chen2025seed} and code generation \cite{seed2025seed, yang2025code, wang2025large}. 
However, formal learning alone primarily optimizes LLMs for predefined tasks, motivating us to draw inspiration from human informal learning to develop broader and more generalizable abilities. 
%However, formal learning alone may not capture broader competencies such as creativity and social reasoning, motivating complementary informal-learning mechanisms.
% However, the broader competencies expected of general-purpose models, including creativity, social reasoning, etc., lie beyond the scope of formal learning and call for learning mechanisms akin to informal learning.

%txy 这里感觉突然扯到了env的事，然后又突然提到了game。task是 game 是不是不太对？playing game/win a game……“Games naturally”就是这里感觉不应该这么开头，换个主语
%This naturally raises a question: what can serve as an informal learning environment for LLMs? We argue that effective environments should: 1) enable learning without explicit instruction, 2) support open-ended interaction through experience and feedback, and 3) generalize across diverse real-world scenarios. Game-based environments naturally satisfy these criteria by offering interactive, feedback-driven sandboxes with intrinsic rewards and broad cognitive diversity. Therefore, games are adopted as a fundamental environment for LLM informal learning.

\begin{figure*}[t]
  \centering
  \includegraphics[width=\textwidth]{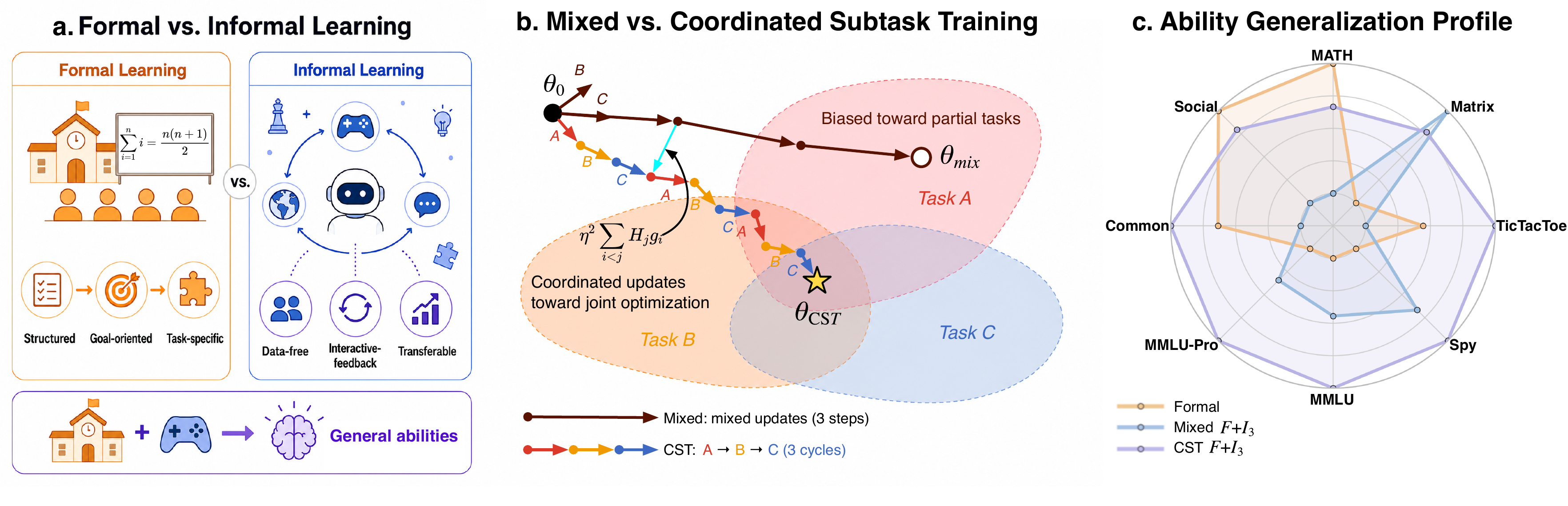}
  \caption{Motivation and framework for informal learning beyond formal task training.
(a) We contrast formal and informal learning as two sources of learning signals.
% We introduce Coordinated Subtask Training (CST), which uses sequential subtask updates to introduce second-order interactions and coordinate multiple subtasks toward a shared objective.
(b) We introduce Coordinated Subtask Training (CST), which uses sequential subtask updates to introduce second-order interactions and coordinate multiple subtasks toward a shared objective. 
(c) 7B-model on the F + $\text{I}_3$ setting results demonstrate improved ability generalization from informal learning and CST.
  }
  \label{fig:nest-intro}
\end{figure*}

This raises a question: what can serve as an informal learning environment for LLMs? We argue that games are a natural choice: they enable learning without explicit instruction, support open-ended interaction through feedback, and abstract diverse real-world scenarios into controllable environments. Building on this insight, we introduce a reinforcement learning framework that jointly trains LLMs on formal mathematical reasoning tasks and informal game-based environments. Specifically, we use three games with increasing interaction complexity: single-turn \textit{Matrix Games}, multi-turn two-player \textit{TicTacToe}, and multi-turn multi-player \textit{Who's the Spy}. Together, they combine formal task structure with informal, experience-driven learning across reasoning, planning, and social interaction.

In this multi-task setting, a straightforward strategy is to mix all subtasks and optimize them with a unified reinforcement learning objective. However, we find that such mixed training suffers from two limitations. At the task-local level, different tasks produce heterogeneous RL signals, including advantages, gradient magnitudes, and exploration dynamics. 
At the cross-task level, multi-task learning should optimize subtasks jointly while avoiding interference among them. However, mixed training simply follows an aggregated gradient, without explicitly encouraging task-specific gradients to be mutually coordinated.

%txy 前面引入太长了 到这个方法之前，有两段差不多了
To address these issues, we propose \textbf{Coordinated Subtask Training (CST)}, which replaces a single mixed update with sequential subtask-specific updates. As shown in Fig.~\ref{fig:nest-intro}, the model is updated on each subtask in turn, with the resulting parameters passed to the next subtask. This decomposes heterogeneous RL signals into cleaner local optimization steps, yielding more faithful subtask-level policy-improvement directions. Beyond local optimization, CST also improves cross-task coordination. Sequential updates introduce second-order interaction terms $\eta^2 H_j g_i$, which we show are equivalent to encouraging larger pairwise gradient inner products across subtasks. Thus, CST implicitly biases optimization toward regions where subtask gradients are better aligned, so updates reinforce rather than interfere with one another. This helps balance in-domain subtask performance with general capabilities under multi-task RL.

Experimental results on ability-oriented benchmarks show that game-based informal learning improves generalization beyond formal learning alone, increasing the average ability score from 38.34\% to 41.13\% for 1.5B models and from 42.00\% to 57.39\% for 7B models. Moreover, CST consistently outperforms mixed training when integrating heterogeneous formal and informal tasks. These results suggest that informal game environments provide useful transferable learning signals, while CST enables them to be absorbed more effectively during multi-task RL training.
Our contributions are summarized as follows:
\begin{itemize}
    %txy 感觉这个点可以写再大一点，而不是落再perspective。可以是个framework或者是个task define。
    \item[1)] \textbf{Game-based informal learning perspective}: We introduce informal learning from educational theory and formulate game as an abstraction of informal learning for LLM training, providing an annotation-free, scalable, and diverse environment;
    %\textbf{Game-based interactive learning}: We formulate game-based environments as a practical abstraction of informal learning for LLM training. This perspective connects educational notions of interaction- and feedback-driven learning with RLVR-style training, showing that games can provide scalable, verifiable, and annotation-free tasks complementary to formal learning;
    %Inspired by human learning processes, we systematically investigate informal learning for LLMs and propose games as an annotation-free environment that naturally supports multi-dimensional ability acquisition;  
    %txy 这个应该直接写方法名，不写Methodological Innovation
    \item[2)]\textbf{Methodological Innovation}: We propose \textbf{CST}, which replaces coarse mixed updates with sequential subtask-specific updates, preserving task-local RL signals and introducing implicit cross-task gradient coordination.  %through second-order interactions between successive subtask updates.
    %We identify the "OR-style" optimization trap in naive mixed training and propose a Nested Training Framework, which transforms the objective into an explicit "AND" logic and ensures the simultaneous acquisition of diverse abilities;
    %We introduce a nested training framework that converts naive multi-task "OR" optimization into an "AND" objective, effectively mitigating task trade-off;
    %这个地方实际并不是mitigate task trade-off，更明显的是提升了多种能力
    %txy 应该是写 xxx sota，或者ood pass mixed之类的。这里的小标题都太概括了
    \item[3)] \textbf{Empirical Validation}: We show that game-based informal learning improves LLM generalization beyond formal learning alone, while CST better integrates heterogeneous multiple tasks by balancing in-domain subtask performance with broader general abilities.
\end{itemize}

\section{Related Works}
\subsection{Games and Large Language Models}
% LLM For Games: Training LLM to master games.
% Game For LLMs: Use game to boost LLM's abilities. Overlook the multiple games settings.
Recent studies on games and LLMs can be broadly categorized into two directions. The first direction, commonly referred as LLM for Games, investigates the training and evaluation of LLMs within specific game environments. Representative works examine LLM behavior in negotiation games, social deduction settings, and multi-agent text-based games, aiming to assess strategic consistency, equilibrium behavior, or task-specific performance \cite{ fan2024can, mao2025alympics, bianchi2024well, guertler2025textarena, akata2025playing}. These studies primarily focus on understanding or improving LLM performance within particular games, rather than enhancing general capabilities across tasks. In contrast, the Games for LLM treats games as structured interaction frameworks for improving broader LLM abilities. Prior works demonstrate that self-play, repeated interactions, and multi-agent game dynamics can facilitate improvements in reasoning, alignment, and strategic adaptation \cite{tang2025game, liu2025spiral, xie2025play}. However, existing studies remain limited in game diversities and provide insufficient analysis on multi-task settings. % different categories of games differentially shape model behavior and capability development.
% Recent benchmarks further systematize this perspective by evaluating LLMs under game-theoretic pressures across diverse strategic settings \cite{ huang2025competing, chen2025llmspark }. 

\subsection{Multi-Task RL Training in LLMs}
Reinforcement learning (RL) has been shown to play a critical role in enhancing the reasoning capabilities of large language models (LLMs) \cite{liu2024deepseek, wang2024reinforcement, khatri2025art, guo2025deepseek, xu2025towards}. A growing body of work demonstrates that RL-based training can significantly improve performance on reasoning-intensive tasks, such as mathematical problem solving \cite{zeng2025simplerl, wang2025reinforcement} and code generation\cite{zhao2025absolute}.More recently, several works have begun to investigate multi-task RL training for LLMs\cite{zeng2025rlve}. %Some studies observe that naively mixing heterogeneous tasks often leads to performance trade-offs\cite{wu2025imbalanced}. To mitigate this issue, 
OMNI-THINKER adopts a curriculum-based training strategy and mixed reward designs \cite{li2025omni} and AgentRL replaces the group-based advantage in GRPO with a task-aware advantage formulation \cite{zhang2025agentrl}.

% 这个地方最好加上和我们的区别，或者不要提这么多？
%In contrast to these approaches, which primarily address multi-task instability through curriculum design, reward shaping, or advantage reweighting, our work explores a complementary direction by reformulating multi-task optimization itself, enabling synergistic ability acquisition without relying on manual designs.
In contrast to these approaches, CST offers a simpler optimization-level alternative: it only changes the update structure, separating subtask-local updates and introducing second-order cross-subtask interactions without manually specification.

% \section{Preliminary}
% In this section, we introduce the conceptual and technical foundations of our approach, including the distinction between formal and non-formal learning, the task formulation and notation used throughout the paper, and the reinforcement learning objectives on which our method is built.

\section{Method}

\begin{figure*}[t]
  \centering
  \includegraphics[width=\textwidth]{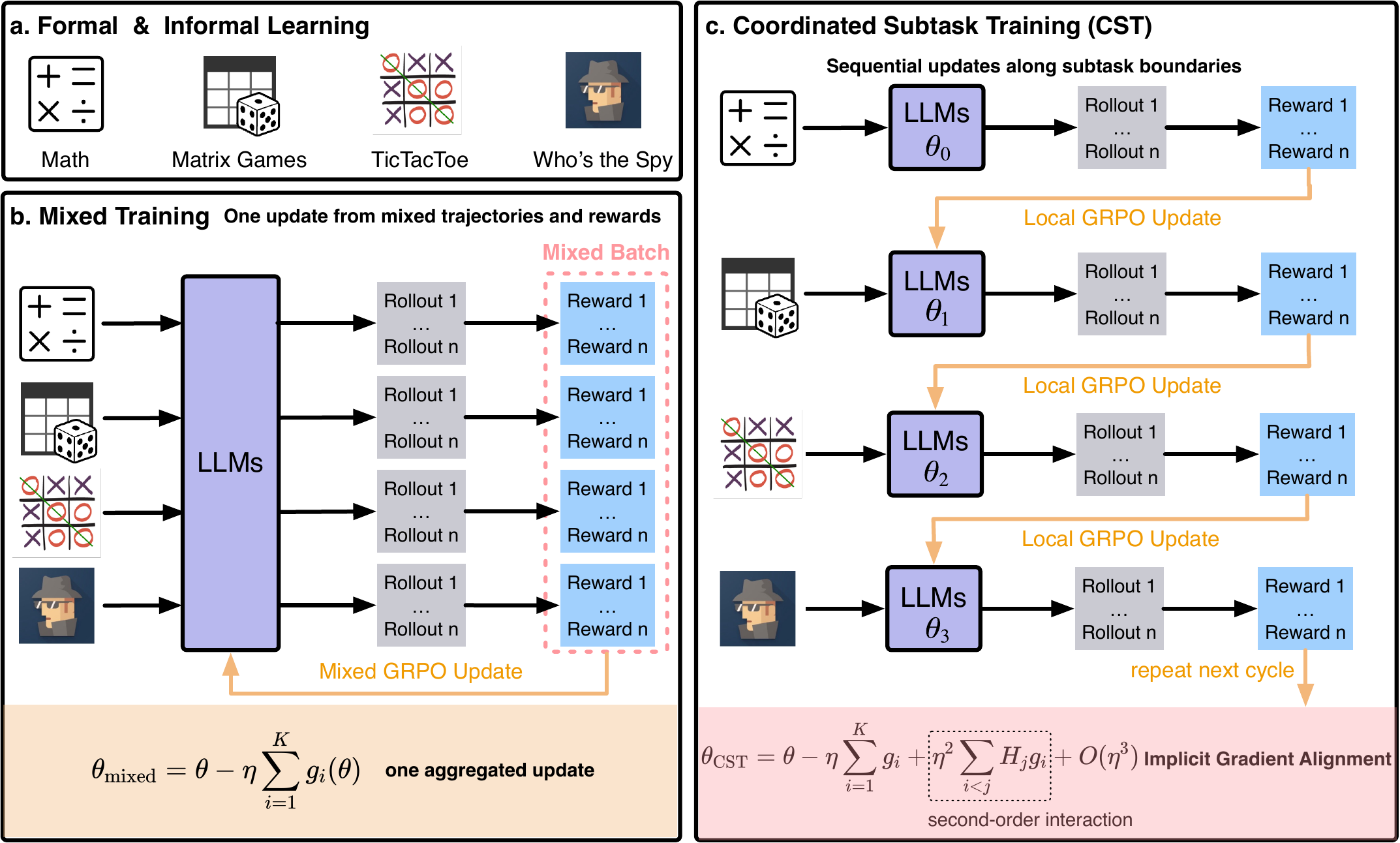}
  \caption{
Overview of our motivation, mixed training and CST framework.
(a) Formal and Informal learning tasks in our trainings. 
(b) Mixed training pools rollouts and rewards from all subtasks into a single GRPO update.
(c) CST instead follows subtask boundaries: each subtask independently generates rollouts and rewards, performs a local GRPO update, and passes the updated parameters to the next subtask in a cyclic order.
  }
  \label{fig:main-architect}
\end{figure*}

In this section, we explore the motivation of introducing formal and informal learning and how informal learning signals can be integrated into model training through reinforcement learning.% , enabling models to balance trade-offs and develop more general and transferable capabilities.

\subsection{Motivation}
\label{sec:learning}
% 介绍认知学基础——显然写的太长应该精简一下

Educational theory distinguishes between \emph{formal} and \emph{informal learning} \cite{coombs1974attacking,johnson2022formal}. Formal learning is organized around structured curricula and predefined objectives, whereas informal learning occurs through everyday interaction, practice, and feedback in open-ended environments \cite{cattell1963theory,horn1967age}. Compared with formal learning, informal learning is often associated with task-independent problem solving and transferable abilities in unseen situations \cite{ziegler2012openness,thorsen2014influence}.

Current LLM training largely follows a formal-learning paradigm, where models are optimized on structured tasks with predefined objectives. This improves task-specific performance but may limit the development of transferable capabilities beyond fixed problem domains.
%Current LLM training largely follows a formal-learning paradigm: models are optimized on structured tasks such as mathematical reasoning and code generation, where objectives and evaluation criteria are explicitly specified. Although effective for improving task-specific performance, this paradigm may bias models toward predefined problem formats and limit the acquisition of broader, transferable capabilities.
Educational research provides further motivation for this direction: informal learning supports problem solving in technology-rich environments \cite{nygren2019lifelong}, contributes substantially to workplace skill development \cite{de2024importance,fevre2001necessary}, and can be especially beneficial when combined with formal learning \cite{gerber2001relationships}. Inspired by these findings, we investigate whether interactive, feedback-driven environments can complement formal task-oriented training and improve the generalizability of LLMs.

\subsection{Game as Informal Learning Environments}

Formal learning emphasizes the structured acquisition of explicit knowledge. In LLM training, this paradigm is naturally instantiated by mathematics, where models solve well-defined problems under supervision. We therefore use \textit{Math} as the formal learning environment in our study.

To model informal learning for LLMs, we ask what can serve as an informal learning environment. We argue that \textsc{game}-based environments are a natural choice: they enable learning without explicit instruction, support open-ended interaction through feedback, and abstract diverse real-world scenarios into controllable environments. Moreover, this makes games suitable for experience-driven learning while retaining clear rules and measurable outcomes for reinforcement learning.

Following this view, we select three games with increasing interaction complexity, including single-turn \textit{Matrix Games},single-turn two-player \textit{TicTacToe}, and multi-turn multi-player \textit{Who's the Spy}, covering different abilities including abstract reasoning, long-horizon planning, and social reasoning. The selected environments are summarized in Table~\ref{tab:game_types}, with Details provided in Appendix~\ref{app:game_rule}.

\subsection{From Mixed to Coordinated Subtask Training}
To combine formal reasoning tasks with game-based informal learning in LLM training, a natural strategy is to train on a mixture of all subtasks. However, our analysis and experiments show that mixed training introduces limitations at both the task-local and cross-task levels. We therefore propose Coordinated Subtask Training (CST) to address these limitations.

\begin{figure}[t]
  \includegraphics[width=\columnwidth]{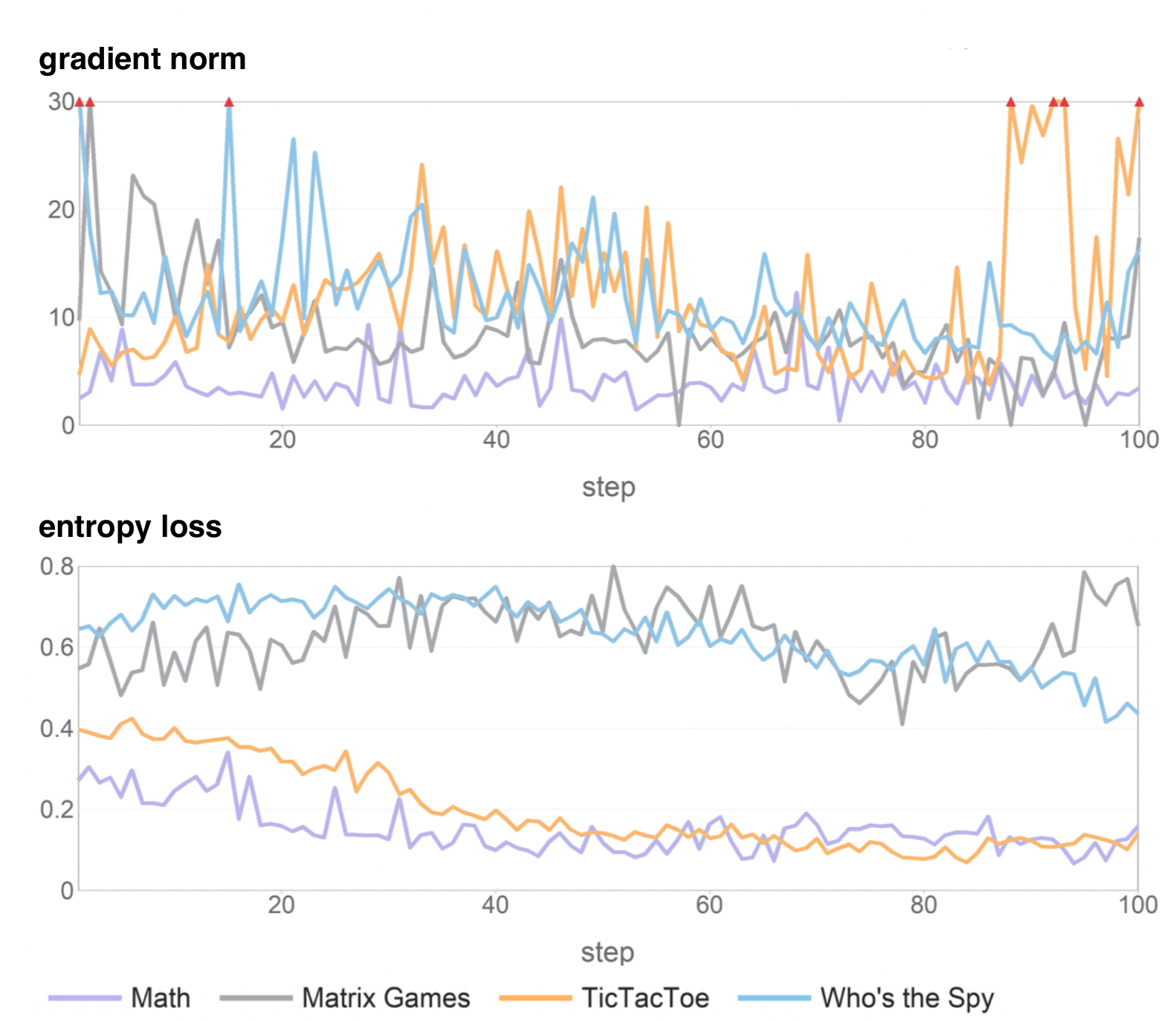}
  \caption{Gradient norms and entropy losses for different subtasks during mixed training.}
  \label{fig:method-mix}
\end{figure}

\subsubsection{Mixed Training}
In this naive mixed training paradigm, trajectories from different subtasks are placed in the same training batch and optimized through a single policy update. 
This design exposes the model to diverse environments, but it forces trajectories with different reward structures, exploration dynamics, and learning stages to share the same RL estimation process.
We argue that this assumption introduces two optimization limitations: \textit{task-local signal contamination} and \textit{cross-task gradient uncoordination}.

Formally, given $K$ subtasks, the mixed update is wrtten as
\begin{equation}
    \theta_{\mathrm{mixed}}
    =
    \theta
    -
    \eta
    \sum_{i=1}^{K} g_i(\theta),
    \label{eq:mixed-update}
\end{equation}
where $g_i(\theta)=\nabla_\theta \mathcal{L}^{(i)}_{\mathrm{RL}}(\theta)$ denotes the policy gradient estimated from subtask $i$.

First, mixed training can cause \textit{task-local signal contamination}. The aggregation in Eq.~\ref{eq:mixed-update} assumes that samples from different subtasks can share the same RL estimation unit. However, subtasks often differ in reward density, exploration demand, trajectory length, and learning stage, leading to distinct gradient scales and entropy patterns, as illustrated in Fig.~\ref{fig:method-mix}. Under shared batch-level calibration, quantities such as advantages, entropy regularization, and policy-gradient scales are estimated across incompatible subtask statistics rather than within each subtask. As a result, the estimated $g_i(\theta)$ can deviate from a clean subtask-level policy-improvement direction.

Second, mixed training suffers from \textit{cross-task gradient uncoordination}. The direct summation in Eq.~\ref{eq:mixed-update} treats multi-task learning as a simple aggregation of subtask gradients, but provides no explicit guidance for coordinating their interactions. Although each $g_i(\theta)$ may still contain task-relevant optimization information, the averaged update $\sum_{i=1}^{K} g_i(\theta)$ does not encourage these directions to become mutually compatible. Conflicting components may therefore interfere within the aggregated update, preventing the model from reaching parameter regions that jointly support multiple subtasks and general capabilities.

\subsubsection{Coordinated Subtask Training}
Therefore, we propose \textbf{Coordinated Subtask Training (CST)}, which improves naive mixed training at both the task-local and cross-task levels. CST replaces the single aggregated update in Eq.~\ref{eq:mixed-update} with a sequence of subtask-specific updates. As shown in Fig.~\ref{fig:main-architect}, each local update uses trajectories from a single subtask, and the resulting parameters are then passed to the next update. This separates heterogeneous RL signals into homogeneous local estimation units, reducing task-local signal contamination. At the same time, sequentially applying subtask updates allows different optimization directions to interact through parameter transitions. We later show that this interaction induces second-order terms that implicitly encourage gradient compatibility across subtasks, addressing cross-task gradient uncoordination.

For clarity, we first consider the basic CST cycle where each subtask performs one local update before switching to the next subtask:
\begin{equation}
    \theta_{\mathrm{CST}}
    =
    U_K
    \circ
    U_{K-1}
    \circ
    \cdots
    \circ
    U_1
    (\theta),
    \label{eq:cst-one-step-cycle}
\end{equation}
where $U_i(\theta)=\theta-\eta g_i(\theta)$.

A second-order expansion around the initial parameter $\theta$ gives
\begin{equation}
    \theta_{\mathrm{CST}}
    =
    \theta
    -
    \eta
    \sum_{i=1}^{K} g_i
    +
    \eta^2
    \sum_{i<j} H_j g_i
    +
    O(\eta^3),
    \label{eq:cst-second-order-basic}
\end{equation}
where $g_i=g_i(\theta)$ and $H_i=\nabla_\theta^2\mathcal{L}^{(i)}_{\mathrm{RL}}(\theta)$ are evaluated at $\theta$.
%Intuitively, they make each later update respond to the parameter changes induced by previous subtasks, instead of treating all subtask gradients as independent components of a single averaged direction.
Following the second-order view of Reptile~\cite{nichol2018reptile}, such terms can be connected to pairwise gradient-alignment objectives under an order-averaged approximation:
\begin{equation}
    \mathbb{E}_{\pi}
    \left[
    \sum_{i<_{\pi}j} H_j g_i
    \right]
    \approx
    \frac{1}{2}
    \nabla_\theta
    \mathbb{E}_{\pi}
    \left[
    \sum_{i<j} g_i^\top g_j
    \right],
    \label{eq:cst-gradient-alignment-basic}
\end{equation}
where $\pi$ denotes the subtask order. 
Therefore, the second-order term in CST acts like an implicit update toward increasing pairwise gradient inner products across subtasks. Since $g_i^\top g_j=\|g_i\|\|g_j\|\cos(g_i,g_j)$, increasing this quantity encourages smaller angles between subtask gradients when their norms are comparable. This biases optimization toward regions where subtask policy-improvement directions are more compatible, allowing progress on different subtasks to reinforce rather than cancel each other.
%This suggests that CST biases optimization toward regions where subtask policy-improvement directions are more compatible. 
% 这里需要说明引入这个额外二阶项的好处
The detailed derivation and discussion about order-averaging condition are provided in Appendix~\ref{app:seq-derivation}.

In practice, CST may allow each subtask to perform $\tau$ consecutive local updates before switching. This block form is written as
\begin{equation}
    \theta_{\mathrm{CST}}
    =
    U_K^{(\tau)}
    \circ
    U_{K-1}^{(\tau)}
    \circ
    \cdots
    \circ
    U_1^{(\tau)}
    (\theta),
    \label{eq:cst-block-cycle}
\end{equation}
where $U_i^{(\tau)}$ denotes applying $U_i$ for $\tau$ consecutive steps. The block length $\tau$ controls the granularity of CST: larger $\tau$ strengthens local consolidation within each homogeneous subtask block, while the same sequential structure preserves the cross-subtask interaction effect above. %A corresponding expansion is given in Appendix~\ref{app:seq-derivation}.

\section{Experiment}

\begin{table*}[t]
\centering
\small
\setlength{\tabcolsep}{3pt}
\renewcommand{\arraystretch}{1.0}

\begin{tabular}{@{}l l c c c c | c c c c c@{}}
% \hline
% \multirow{2}{*}{Setting}
% & \multirow{2}{*}{Model}
% & \multicolumn{4}{c}{\textbf{In-Domain Tasks}}
% & \multicolumn{4}{c}{\textbf{Out-of-Distribution Tasks}} \\
% \cmidrule(lr){3-6} \cmidrule(lr){7-10}
% & 
\toprule
\makecell[c]{\textbf{Setting}}
& \makecell[c]{\textbf{Model}}
& \makecell[c]{\textbf{MATH}}
& \makecell[c]{\textbf{Matrix}}
& \makecell[c]{\textbf{TicTacToe}}
& \makecell[c]{\textbf{Spy}}
& \makecell[c]{\textbf{MMLU}}
& \makecell[c]{\textbf{MMLU-Pro}}
& \makecell[c]{\textbf{Common}}
& \makecell[c]{\textbf{Social}} 
& \makecell[c]{\textbf{Avg.}}\\
\midrule

Base
& Qwen2.5-1.5B
& 17.20 & 7.00 & 1.00 & 2.00 & 37.87 & 13.49 & 7.79 & 27.64 & 27.10 \\
\midrule
Formal
& \textit{Math}
& \cellcolor{grey_purple} 43.20 & 21.00 & 4.00 & 9.00 & 51.38 & 20.97 & 15.08 & 65.92 & 38.34 \\
\midrule

\multirow{3}{*}{Informal}
& \textit{Matrix Games}
& 19.20 & \cellcolor{grey_purple} 44.00 & 34.00 & 14.00 & 43.79 & 18.00 & 9.55 & 64.12 & 33.87 \\
& \textit{TicTacToe}
& 22.40 & 32.00 & \cellcolor{grey_purple} 75.00 & 21.00 & 47.89 & 16.45 & 11.56 & 64.12 & 35.00 \\
& \textit{Who's the Spy}
& 19.20 & 21.00 & 0.00 & \cellcolor{grey_purple} 33.00 & 43.85 & 15.68 & 20.85 & 60.70 & 35.27 \\
\midrule

\multirow{2}{*}{F + $\text{I}_1$}
& mixed
& \cellcolor{grey_purple} 37.80 & \cellcolor{grey_purple} 29.00 & 0.00 & 16.00 & 45.19 & 17.00 & 12.81 & 59.11 & 33.53 \\
& \textbf{CST}
& \cellcolor{grey_purple} 38.00 & \cellcolor{grey_purple} 49.00 & 11.00 & 20.00 & \textbf{52.21} & \textbf{22.90} & \textbf{16.58} & \textbf{64.23} & \textbf{38.98} \\
% & nested
% & \cellcolor{grey_purple} 40.00 & \cellcolor{grey_purple} 65.00 & 8.00 & 24.00 & \textbf{52.94} & \textbf{21.03} & \textbf{21.11} & \textbf{66.53} & \textbf{40.40} \\
\midrule

\multirow{2}{*}{F + $\text{I}_2$}
& mixed
& \cellcolor{grey_purple} 28.00 & \cellcolor{grey_purple} 30.00 & \cellcolor{grey_purple} 50.00 & 20.00 & 51.35 & 19.47 & 17.84 & 65.30 & 38.49 \\
& \textbf{CST}
& \cellcolor{grey_purple} 36.80 & \cellcolor{grey_purple} 65.00 & \cellcolor{grey_purple} 41.00 & 31.00 & \textbf{52.35} &  \textbf{20.10} &  \textbf{24.37} &  \textbf{67.76} &  \textbf{41.13} \\
% & nested
% & \cellcolor{grey_purple} 16.20 & \cellcolor{grey_purple} 20.00 & \cellcolor{grey_purple} 16.00 & 26.00 & \textbf{53.08} & \textbf{20.20} & \textbf{20.85} & \textbf{67.09} & \textbf{40.31} \\
\midrule

\multirow{2}{*}{F + $\text{I}_3$}
& mixed
& \cellcolor{grey_purple} 35.80 & \cellcolor{grey_purple} 44.00 & \cellcolor{grey_purple} 28.00 & \cellcolor{grey_purple} 17.00 & \textbf{50.37} & \textbf{20.92} & 17.09 & 65.81 & 38.55 \\
& \textbf{CST}
& \cellcolor{grey_purple} 31.80 & \cellcolor{grey_purple} 55.00 & \cellcolor{grey_purple} 80.00 & \cellcolor{grey_purple} 41.00 & 49.47 & 18.39 & \textbf{28.89} & \textbf{66.58} & \textbf{40.83} \\
\bottomrule

\end{tabular}

\caption{Performance of single-tasks, mixed multi-tasks, and CST multi-tasks training for base model Qwen2.5-1.5B-Instruct, where F + $\text{I}_k$ denotes the combination of formal learning and $k$ informal learning tasks. %We use MATH, Matrix, Spy, Common and Social to denote MATH500, \textit{Matrix Games}, \textit{Who's the Spy}, CommonGen and SocialIQA respectively. Avg represents the average score of general abilities, including MMLU, MMLU-Pro, CommonGen and SocialIQA. 
Purple shading highlights the in-domain tasks for each setting.}% Purple shading highlights the improvement of nested training. }
\label{tab:main_results_1.5B}
\end{table*}

% -------------- 7B models -1 ---------------
% 这个{table*}[t]可能就是要分散在两页里，可能需要把7B结果放在前面？到时候看一下
\begin{table*}[t]
\centering
\small
\setlength{\tabcolsep}{3pt}
\renewcommand{\arraystretch}{1.0}

\begin{tabular}{@{}l l c c c c | c c c c c@{}}
% \hline
% \multirow{2}{*}{Setting}
% & \multirow{2}{*}{Model}
% & \multicolumn{4}{c}{\textbf{In-Domain Tasks}}
% & \multicolumn{4}{c}{\textbf{Out-of-Distribution Tasks}} \\
% \cmidrule(lr){3-6} \cmidrule(lr){7-10}
% & 
\toprule
\makecell[c]{\textbf{Setting}}
& \makecell[c]{\textbf{Model}}
& \makecell[c]{\textbf{MATH}}
& \makecell[c]{\textbf{Matrix}}
& \makecell[c]{\textbf{TicTacToe}}
& \makecell[c]{\textbf{Spy}}
& \makecell[c]{\textbf{MMLU}}
& \makecell[c]{\textbf{MMLU-Pro}}
& \makecell[c]{\textbf{Common}}
& \makecell[c]{\textbf{Social}} 
& \makecell[c]{\textbf{Avg.}}\\
\midrule
Base
& Qwen2.5-7B
& 54.60 & 40.00 & 40.00 & 25.00
& 71.43 & 40.17 & 26.38 & 75.44 & 53.36 \\
\midrule
Formal
& \textit{Math}
& \cellcolor{grey_purple} 58.40 & 39.00 & 31.00 & 18.00 
& 66.51 & 32.45 & 35.18 & 75.90 & 42.00 \\
\midrule

\multirow{3}{*}{Informal}
& \textit{Matrix Games}
& 58.20 & \cellcolor{grey_purple} 64.00 & 23.00 & 27.00 
& 73.22 & 44.51 & 26.13 & 76.20 & 55.02 \\
& \textit{TicTacToe}
& 49.80 & 57.00 & \cellcolor{grey_purple} 78.00 & 33.00 
& 73.37 & 46.35 & 28.89 & 76.41 & 56.26 \\
& \textit{Who's the Spy}
& 55.60 & 43.00 & 38.00 & \cellcolor{grey_purple} 37.00 
& 71.33 & 41.17 & 28.39 & 74.92 & 53.95 \\
\midrule

\multirow{2}{*}{F + $\text{I}_1$}
& mixed
& \cellcolor{grey_purple} 58.20 & \cellcolor{grey_purple} 52.00 & 24.00 & 21.00 
& 67.55 & 35.21 & 27.89 & 73.34 & 51.00\\
& \textbf{CST}
& \cellcolor{grey_purple} 59.40 & \cellcolor{grey_purple} 44.00 & 32.00 & 35.00 
& \textbf{68.45} & \textbf{39.01} & \textbf{35.43} & \textbf{74.82} & \textbf{54.43} \\
\midrule

\multirow{2}{*}{F + $\text{I}_2$}
& mixed
& \cellcolor{grey_purple} 50.00 & \cellcolor{grey_purple} 63.00 & \cellcolor{grey_purple} 45.00 & 9.00 
& 66.80 & 37.81 & 36.68 & 72.21 & 53.38 \\
& \textbf{CST}
% | 56.20   | 51.00 | 55.00     | 30.00      | 68.00  | 35.04    | 46.73  | 74.92      | 56.17 |
& \cellcolor{grey_purple} 56.20 & \cellcolor{grey_purple} 51.00 & \cellcolor{grey_purple} 55.00 & 30.00 
& \textbf{68.00} & \textbf{35.04} & \textbf{46.73} & \textbf{74.92} & \textbf{56.17}\\
\midrule

\multirow{2}{*}{F + $\text{I}_3$}
% | 56.60   | 52.00 | 27.00     | 34.00      | 69.24  | 36.44    | 31.66  | 74.62      | 52.99 |
& mixed
& \cellcolor{grey_purple} 56.60 & \cellcolor{grey_purple} 52.00 & \cellcolor{grey_purple} 27.00 & \cellcolor{grey_purple} 34.00 
& 69.24 & 36.44 & 31.66 & 74.62 & 52.99 \\
& \textbf{CST}
& \cellcolor{grey_purple} 57.80 & \cellcolor{grey_purple} 49.00 & \cellcolor{grey_purple} 36.00 & \cellcolor{grey_purple} 42.00 
& \textbf{72.65} & \textbf{44.08} & \textbf{37.19} & \textbf{75.64} & \textbf{57.39} \\
\bottomrule

\end{tabular}

\caption{Performance of single-tasks, mixed multi-tasks, and CST multi-tasks training for base model Qwen2.5-7B-Instruct, where F + $\text{I}_k$ denotes the combination of formal learning and $k$ informal learning tasks.}
\label{tab:main_results_7B}
\end{table*}

In this section, we present experimental results to evaluate the effectiveness of combining formal and informal learning, as well as the impact of the proposed CST framework.

\subsection{Environment Setup}
We briefly describe the environments, training configurations, and evaluation protocols used in our experiments, with details in the Appendix.

\subsubsection{Environments and Tasks}
We adopt \textit{Math} as the formal learning environment and a set of game-based environments as informal learning tasks, including \textit{Matrix Games}, \textit{TicTacToe}, and \textit{Who's the Spy}. For \textit{Math} training, we use MathLv3-5 problems from the SimpleRL-Zoo-Data \cite{zeng2025simplerl}. For game-based training, we refer to Appendix~\ref{app:game_rule} for detailed environment descriptions and prompting strategies.

\subsubsection{Training Settings}
Following the RAGEN framework, we train language models using $\text{StarPO}^*$, with trajectory-based reinforcement learning formulation using GRPO. We evaluate two model scales, Qwen2.5-1.5B-Instruct and Qwen2.5-7B-Instruct \cite{qwen2,qwen2.5}, as base models. In two-player and multi-player games, we employ Qwen3-14B as opponents \cite{qwen3technicalreport}. 
For CST, we set the block length to $\tau=4$ for Qwen2.5-1.5B and $\tau=1$ for Qwen2.5-7B. We use a larger block length for the smaller model to provide stronger task-local consolidation, while the larger model adopts the basic one-step CST cycle.
Full training parameters and implementation are reported in Appendix~\ref{app:experiment}.

We report three main experimental settings: single-task training, multi-task mixed training, and multi-task CST. For multi-task configurations, we investigate progressive combinations of formal and informal learning and denote the setups as $\text{F + I}_k$, where $k$ indicates the number of informal learning components included in training, reflecting the depth and complexity of informal learning. Specifically, $\text{I}_1$, $\text{I}_2$, and $\text{I}_3$ correspond to \textit{Matrix Games}, \textit{Matrix Games} combined with \textit{TicTacToe}, and the full combination including \textit{Matrix Games}, \textit{TicTacToe}, and \textit{Who's the Spy}, respectively.

\begin{table*}[t]
\centering
\small
\setlength{\tabcolsep}{3pt}
\renewcommand{\arraystretch}{1.0}
\begin{tabular}{@{}l l c c c c | c c c c c@{}}
\toprule
\makecell[c]{\textbf{Setting}}
& \makecell[c]{\textbf{Model}}
& \makecell[c]{\textbf{MATH}}
& \makecell[c]{\textbf{Matrix}}
& \makecell[c]{\textbf{TicTacToe}}
& \makecell[c]{\textbf{Spy}}
& \makecell[c]{\textbf{MMLU}}
& \makecell[c]{\textbf{MMLU-Pro}}
& \makecell[c]{\textbf{Common}}
& \makecell[c]{\textbf{Social}} 
& \makecell[c]{\textbf{Avg.}}\\
\midrule

\multirow{2}{*}{F + $\text{I}_1$}
% & mixed
% &\cellcolor{grey_purple} 37.80 & \cellcolor{grey_purple} 29.00 & 0.00 & 16.00 & 45.19 & 17.00 & 12.81 & 59.11 & 33.53 \\
& TaskAdv
& \cellcolor{grey_purple} 30.40 & \cellcolor{grey_purple} 65.00 & 10.00 & 33.00 & 45.89 & 18.52 & \textbf{18.34} & 56.55 & 34.83 \\
& \textbf{CST}
% | 38.00 | 49.00  | 11.00     | 20.00 | 52.21 | 22.90    | 16.58  | 64.23  | 38.98 |
& \cellcolor{grey_purple} 38.00 & \cellcolor{grey_purple} 49.00 & 11.00 & 20.00 & \textbf{52.21} & \textbf{22.90} & 16.58 & \textbf{64.23} & \textbf{38.98} \\
\midrule

\multirow{2}{*}{F + $\text{I}_2$}
% & mixed
% & \cellcolor{grey_purple} 28.00 & \cellcolor{grey_purple} 30.00 & \cellcolor{grey_purple} 50.00 & 20.00 & 51.35 & 19.47 & 17.84 & 65.30 & 38.49 \\
& TaskAdv
& \cellcolor{grey_purple} 10.40 & \cellcolor{grey_purple} 35.00 & \cellcolor{grey_purple} 46.00 & 18.00 & 47.20 & 17.64 & 22.11 & 66.02 & 38.24 \\
& \textbf{CST}
& \cellcolor{grey_purple} 36.80 & \cellcolor{grey_purple} 65.00 & \cellcolor{grey_purple} 41.00 & 31.00 & \textbf{52.35} & \textbf{20.10} & \textbf{24.37} & \textbf{67.76} & \textbf{41.13} \\

\bottomrule
\end{tabular}
\caption{Comparison between mixed training, task-level normalization, and CST with 1.5B base model.}
\label{tab:task_norm}
\end{table*}

\subsubsection{Evaluation Metrics}
We evaluate in-domain performance on the training tasks, including the MATH500 benchmark \cite{lightman2023lets}, \textit{Matrix Games}, \textit{TicTacToe}, and \textit{Who’s the Spy}. For games with opponents, including \textit{TicTacToe} and \textit{Who's the Spy}, we report the average success rate against Gemini-2.5-Flash in 100 rounds \cite{comanici2025gemini}. For general and diverse abilities, we choose MMLU and MMLU-Pro for multi-domain reasoning \cite{hendrycks2021ethics,hendryckstest2021}, CommonGen for creative language generation \cite{lin-etal-2020-commongen} and SocialIQA for social reasoning \cite{sap2019socialIQa}. In CommonGen tasks, we use GPT-4o \cite{hurst2024gpt} to compare the generated outputs against ground truth references. %,  counting better and semantically equivalent generations (ties) as success. %All reported results are percentages, and the percentage symbol (\%) is omitted for brevity.
The details are in Appendix~\ref{app:common}.

\paragraph{Notation.}
We abbreviate MATH500, \textit{Matrix Games}, \textit{Who's the Spy}, CommonGen, and SocialIQA as MATH, Matrix, Spy, Common, and Social. Avg. denotes the mean score over MMLU, MMLU-Pro, CommonGen, and SocialIQA. All results are reported as percentages, with the percentage symbol omitted. 

\subsection{Main Results}

% \begin{figure}[t]
%   \includegraphics[width=\columnwidth]{image/mix-nest.pdf}
%   \caption{Comparison between mixed and nested training in F + $\text{I}_2$ setting with 7B base model. Vertical axis denotes the performance, purple color denotes the ID tasks and gray color denotes the general abilities.}
%   \label{fig:stable-train}
% \end{figure}

Table~\ref{tab:main_results_1.5B} and~\ref{tab:main_results_7B} summarize the main results across two model scales. We analyze them from two perspectives: the role of formal versus informal learning and the effectiveness of CST method.

\paragraph{Formal vs. Informal Learning.}
We first examine whether introducing informal learning is empirically meaningful. The results on single-task settings show that formal and informal learning exhibit complementary effects across model scales. For the 1.5B model, formal math training brings most gains on reasoning-oriented benchmarks, indicating the value of structured supervision for smaller models. Meanwhile, informal tasks provide additional benefits by introducing interactive and context-dependent signals. 
For the 7B model, informal learning yields stronger overall improvements, likely because game-based informal environments introduce less familiar interaction patterns and feedback structures.
For instance, \textit{Who's the Spy} promotes contextual description, which aligns with open-ended generation tasks such as CommonGen. These results indicate that formal and informal learning enhance different aspects of model capability. 

\paragraph{Effectiveness of CST.}
Although both formal and informal learning are beneficial when applied in isolation, naively mixing their learning signals does not consistently translate into performance gains.
In contrast, CST substantially alleviates performance degradation on general abilites across \textbf{all} settings.
The effect is particularly pronounced in the $\text{F + I}_1$ setting with 1.5B base model and $\text{F + I}_3$ setting with 7B base model, where CST yields \textbf{5.45\% and 4.4\% absolute improvement}.

\begin{figure}[t]
  \includegraphics[width=\columnwidth]{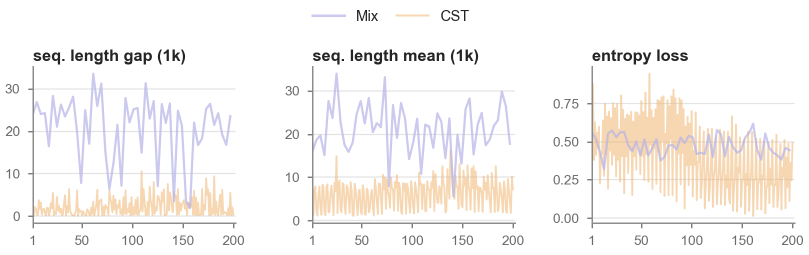}
  \caption{Training dynamics of mixed training and CST in F + $\text{I}_3$ setting with Qwen2.5-7B.}
  \label{fig:exp-metric}
\end{figure}

To further analyze the effectiveness, we compare the training dynamics of mixed training and CST in Fig.~\ref{fig:exp-metric}, where one mixed step is aligned with four consecutive CST subtask updates. Mixed training shows larger sequence-length gaps and higher average sequence lengths, indicating stronger within-update heterogeneity across subtasks. CST reduces these discrepancies by using subtask-specific local updates, yielding cleaner and more homogeneous RL estimation units. The entropy-loss curve further suggests that CST maintains active exploration dynamics while avoiding the large sequence heterogeneity observed in mixed training. %These results provide empirical evidence that CST mitigates the local signal heterogeneity introduced by naive mixed updates.

\subsection{Case Study}

\begin{table*}[t]
\centering
\small
\setlength{\tabcolsep}{3pt}
\renewcommand{\arraystretch}{1.0}
\begin{tabular}{@{}l l c c c c | c c c c c@{}}
% \hline
% \multirow{2}{*}{Setting}
% & \multirow{2}{*}{Model}
% & \multicolumn{4}{c}{\textbf{In-Domain Tasks}}
% & \multicolumn{4}{c}{\textbf{Out-of-Distribution Tasks}} \\
% \cmidrule(lr){3-6} \cmidrule(lr){7-10}
% & 
\toprule
\makecell[c]{\textbf{Setting}}
& \makecell[c]{\textbf{Model}}
& \makecell[c]{\textbf{MATH}}
& \makecell[c]{\textbf{Matrix}}
& \makecell[c]{\textbf{TicTacToe}}
& \makecell[c]{\textbf{Spy}}
& \makecell[c]{\textbf{MMLU}}
& \makecell[c]{\textbf{MMLU-Pro}}
& \makecell[c]{\textbf{Common}}
& \makecell[c]{\textbf{Social}} 
& \makecell[c]{\textbf{Avg.}}\\
\midrule

\multirow{2}{*}{F + $\text{I}_2$}
& mixed
& \cellcolor{grey_purple} 28.00 & \cellcolor{grey_purple} 30.00 & \cellcolor{grey_purple} 50.00 & 20.00 & 51.35 & 19.47 & 17.84 & 65.30 & 38.49 \\
& \textbf{CST}
& \cellcolor{grey_purple} 36.80 & \cellcolor{grey_purple} 65.00 & \cellcolor{grey_purple} 41.00 & 31.00 & \textbf{52.35} &  \textbf{20.10} &  \textbf{24.37} &  \textbf{67.76} &  \textbf{41.13} \\
\midrule

\multirow{2}{*}{$\text{I}_2$}
& mixed & 9.40 & \cellcolor{grey_purple} 49.00 & \cellcolor{grey_purple} 51.00 & 26.00 & 39.57 & 15.78 & 17.34 & 59.98 & 33.17 \\
& \textbf{CST} & 10.00 & \cellcolor{grey_purple} 50.00 & \cellcolor{grey_purple} 50.00 & 37.00 &  \textbf{41.16} &  \textbf{15.97} &  \textbf{24.37} &  \textbf{60.39} &  \textbf{35.47} \\

\bottomrule
\end{tabular}
\caption{Effect of formal learning and CST under the informal-only setting $\text{I}_2$ with 1.5B base model.}
\label{tab:ablation_informal}
\end{table*}

\begin{figure}[t]
  \includegraphics[width=\columnwidth]{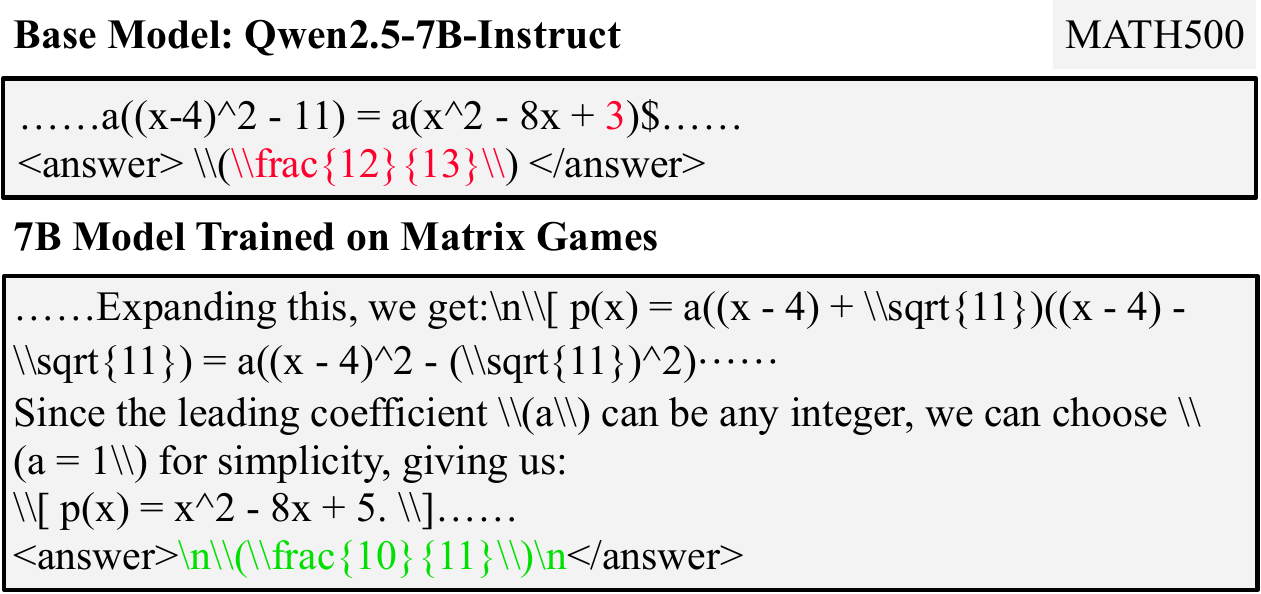}
  \caption{Case Studies on MATH500 benchmark.}
  %\caption{Cases of MATH500 answers between base and model after \textit{Matrix Games}, and CommonGen answers between mixed and nested training 7B models in $\text{F + I}_2$ setting.}
  \label{fig:case-mini}
\end{figure}

To illustrate the effectiveness of informal learning, we present a compact case study in Fig.~\ref{fig:case-mini} illustrating results of MATH500, with full and detailed comparisons in the Appendix (Fig.~\ref{fig:case-matrix-math} and~\ref{fig:case-mnc-common}). 
On the MATH500 benchmark, the base 7B model follows the correct high-level reasoning but makes a subtle arithmetic error, leading to an incorrect result, whereas the \textit{Matrix Games}-trained model maintains a more explicit and verifiable derivation. Moreover, it  chooses $a=1$ creatively to simplify the calculation.
On the CommonGen benchmark, while mixed training on $\text{F + I}_2$ setting produces a reasonable sentence, CST encourages deeper integration by self-validating the result. 
Together, these cases demonstrate that informal learning with the CST framework promotes more explicit, robust, and creative reasoning across both mathematical and generative tasks.

\subsection{Ablation Study}
% 大量消融实验，这个地方可以放几个问句那种
%We conduct a series of ablation studies to further analyze the generality, necessity, and sensitivity of the proposed CST framework. 
% 大量消融实验，这个地方可以放几个问句那种
We conduct ablation studies to analyze CST under multi-task RL, comparing it with task-level advantage normalization, assessing the necessity of informal learning, and evaluating its generality across formal-informal task combinations. The analysis of opponent sensitivity is provided in Appendx~\ref{app:opp}.

\paragraph{comparison with Task-Level Advantage.}
We compare CST with TaskAdv, the task-level advantage normalization strategy proposed in AgentRL~\cite{zhang2025agentrl}. TaskAdv normalizes advantages within each task, partially addressing task-local imbalance, but still optimizes through a single mixed update.
In contrast, CST uses separate subtask updates, yielding cleaner local signals and second-order interactions that implicitly promote cross-task gradient alignment. As shown in Table~\ref{tab:task_norm}, CST achieves a better balance between in-domain subtask performance and broader general abilities than TaskAdv.

\paragraph{Necessity of Formal vs Informal Learning}
The main experiments already show that combining formal and informal learning yields stronger general abilities than formal learning alone. Here, we further examine whether informal learning is sufficient, as assumed in many game-centric LLM training approaches. Using the same $\text{I}_2$ configuration on the 1.5B base model, we compare informal-only training against the formal+informal($\text{F + I}_2$) setting. As shown in Table~\ref{tab:ablation_informal}, models trained solely on informal learning tasks consistently underperform those trained with both formal and informal learning across most benchmarks, under both mixed training and CST. In particular, removing \textit{Math} signals in CST leads to a performance drop of 11.19\% on MMLU and 5.66\% on the average general ability score. These results highlight the critical role of formal learning, suggesting that it cannot be fully replaced by informal learning alone.

\paragraph{Generality of CST.}
To examine whether the CST framework generalizes beyond the combination of formal and informal learning, we evaluate it under an informal-learning-only setting, denoted as $\text{I}_2$, which consists of \textit{Matrix Games} and \textit{TicTacToe}. As shown in Table~\ref{tab:ablation_informal}, CST remains effective in this setting, improving CommonGen performance from 17.34\% to 24.37\% and increasing the average general ability score from 33.17\% to 35.47\%. These results demonstrate that the CST framework provides stable gains even when applied solely to informal learning environments, indicating its robustness beyond the formal-informal combination.

\section{Conclusion}

Inspired by theories in cognitive science, we model formal learning as structured \textit{Math} reasoning tasks and informal learning as interactive, game-based environments. We design three representative game settings: \textit{Matrix Games}, \textit{TicTacToe}, and \textit{Who's the Spy}. While a naive mixed training strategy leads to degradation at both task-local and cross-task level, we show that the CST framework enables effective task-local signal isolation and cross-task gradient coordination. Extensive experiments demonstrate that the combination of formal and informal learning is necessary. Compared to mixed training, CST improves generalization across diverse settings consistently, resulting in LLMs with broader abilities.

\section{Limitations}

This work considers a limited set of formal and informal learning environments, and the game designs represent only a subset of possible interactive settings. While the CST framework generalizes beyond the studied combinations, its effectiveness in more complex or real-world interactive environments remains to be explored. Moreover, our experiments use fixed opponent models in multi-agent games, and extending to self-evolve settings or adaptive opponents is left for future work.

\bibliography{custom}

\appendix

% \section{Results of Naively Mixing}
% \label{sec:exp_mix}
% 这个地方可以用图表的形式，折线图之类的展示mix训练不稳定的问题

\section{Game Rules}
\label{app:game_rule}
We categorize games into three types according to their interaction structure: single-turn games, multi-turn two-player games, and multi-turn multi-player games. For each category, we select a representative game, namely \textit{Matrix Games}, \textit{TicTacToe}, and \textit{Who's the Spy}, respectively. Different games cover different reasoning abilities, as shown in Table.~\ref{tab:game_types}.

Following the terminology in VeRL~\cite{sheng2024hybridflow}, we adopt the concept of \textbf{multi-turn} rather than \textbf{multi-step}. Here, a \textbf{turn} is defined as one complete interaction round in which the trained LLM is queried to produce an action or decision, contributing to the overall game trajectory, while multi-turn refers to a game trajectory with multiple turns.
For example, \textit{Who's the Spy} is considered a three-turn game, as each player, as well as the trained LLM,  participates in three major interaction rounds: two description turns and one final voting turn. Each turn corresponds to a distinct LLM query, and the sequence of these turns together constitutes a full trajectory. 

\begin{table*}[t]
\centering
\small
\renewcommand{\arraystretch}{1.0}
\begin{tabular}{l l l}
\hline
\textbf{Category} & \textbf{Game} & \textbf{Targeted Abilities} \\
\hline
Single-turn 
& \textit{Matrix Games} % (Nash Equilibrium) 
& Abstract \& Strategic reasoning; \\

Multi-turn two-player 
& \textit{TicTacToe}
& Long-horizon planning; Sequential decision making; \\

Multi-turn multi-player 
& \textit{Who's the Spy} 
& Theory of Mind; Creative language generation \\
\hline
\end{tabular}
\caption{Representative game environments and the reasoning abilities they promote.}
\label{tab:game_types}
\end{table*}

\subsection{Matrix Games}
\begin{figure*}[t]
  \centering
  \includegraphics[width=\textwidth]{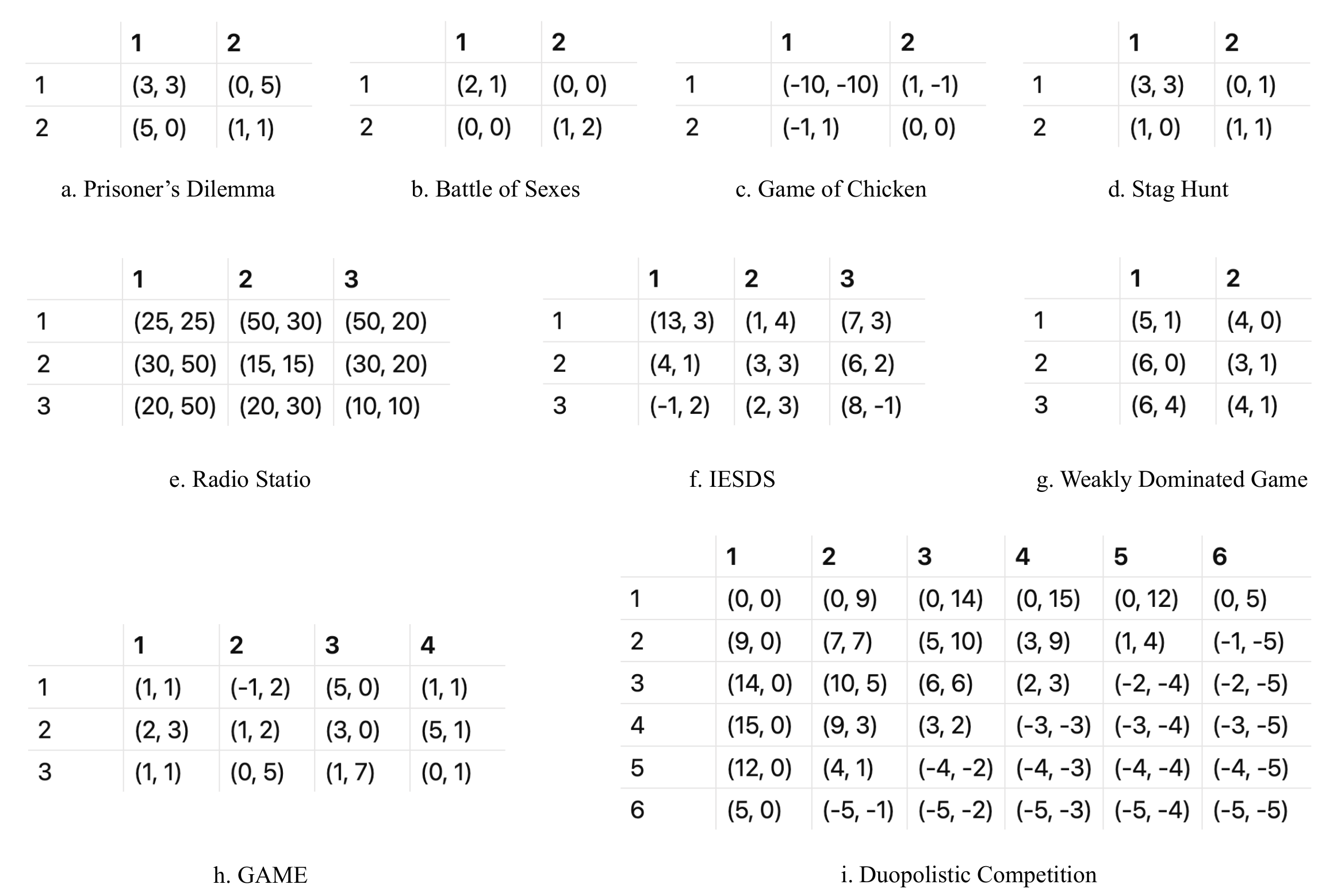}
  \caption{Detailed payoff matrices used in matrix game environments.}
  \label{fig:matrix}
\end{figure*}
\textit{Matrix Games} are single-turn strategic reasoning games. We adopt a set of classic matrix games to train LLMs in strategic reasoning, specifically their ability to infer opponents' actions and optimize their own policies toward Nash Equilibrium. In these games, the model must select an action based solely on an abstract payoff table, without access to domain-specific semantics or external knowledge. This setting encourages \emph{abstract reasoning} by requiring the model to interpret symbolic payoffs, compare outcomes across action pairs, and reason about best responses under different opponent choices. At the same time, it fosters \emph{strategic reasoning}: since rewards depend on both players' decisions, the model must anticipate the opponent's likely move, consider mutual incentives, and choose actions that remain robust under strategic interaction, rather than maximizing immediate payoff in isolation. Through reinforcement learning over repeated interactions, the model is trained to align its action selection with equilibrium-consistent behavior, learning to balance self-interest with opponent-aware reasoning. 

Following \cite{hua2024game}, we select a collection of well-studied matrix games that are verified to admit Nash Equilibrium solutions, including Prisoner's Dilemma, Battle of the Sexes, Game of Chicken, Stag Hunt, Radio Station, IESDS, Duopolistic Game, GAME, and Weakly Dominated Game. The corresponding payoff matrices are illustrated in Fig.~\ref{fig:matrix}. 
To prevent the model from overfitting to a fixed numerical scale or specific prompt format, we apply random transformations to the payoff matrices during training. Specifically, each matrix is randomly multiplied by $-1$ or left unchanged, and an offset of $\pm 100$ is added to all entries. These transformations preserve the strategic structure and equilibrium properties of the games while encouraging the model to focus on relative payoffs and strategic relationships rather than absolute values.
In addition, we design multiple instruction prompt templates to present matrix games under diverse linguistic and contextual formulations. This further improves robustness to prompt variations and discourages reliance on superficial patterns. An illustrative example of such a prompt template is shown below:

\begin{promptbox}
{role}.

Rows = Player 1's actions [{p1_actions_list}]; Columns = Player 2's actions [{p2_actions_list}].

### P1's payoff
{p1_payoff_table}

### P2's payoff
{p2_payoff_table}

{instr}
\end{promptbox}

% \begin{lstlisting}[basicstyle=\ttfamily\footnotesize,breaklines=true]
% Rows = Player 1's actions [{p1_actions_list}]; Columns = Player 2's actions [{p2_actions_list}].
% \end{lstlisting}

\subsection{TicTacToe}
\textit{TicTacToe} is a multi-turn two-player board game in which players alternately place symbols (e.g., \texttt{X} and \texttt{O}) on a $3 \times 3$ grid, with the objective of forming a straight line of three identical symbols horizontally, vertically, or diagonally. Successful play requires each player to anticipate the opponent's potential actions and strategically plan subsequent moves, thereby encouraging the model to reason about others' intentions and future behaviors. In addition, \textit{TicTacToe} imposes strict constraints on valid actions: only unoccupied grid positions can be selected at each turn. These explicit legality requirements provide a clear supervision signal for action validity, which further strengthens the model's instruction-following and rule-compliance capabilities. Prior cognitive and educational studies have also shown that turn-based board games like \textit{TicTacToe} are effective for training planning and strategic reasoning skills \cite{noda2019effectiveness}, supporting its suitability as an informal learning environment.

To enhance training robustness and reduce prompt-specific bias, we design four distinct initial prompts that describe the game setting and interaction protocol from different perspectives. Moreover, we observe that the win conditions are not always trivially recognized by LLMs, especially in multi-turn settings. To address this issue, we further introduce an auxiliary win-condition prompt that explicitly explains the winning criteria, including illustrative examples such as horizontal, vertical, and diagonal line completions. During training, the initial prompt is randomly sampled from the prompt set, and the win-condition prompt is also randomly included to improve generalization. Qwen3-14B is employed as the opponent model during training, considering its strong overall capabilities and efficient local inference speed. An illustrative example of the prompt template with the win-condition description is provided below:

\begin{promptbox}
##Game Rules: TicTacToe
**Objective**: Be the first player to connect 3 of your pieces in a continuous line.
**Player Pieces**:
- Player 1: 'O'
- Player 2: 'X'
- Empty Slot: '.'
**How to Play**:
1. The game is played on a 3x3 vertical grid.
2. Players take turns setting one of their pieces into any available slot.
**Winning Conditions**:
The game ends when a player forms a line of 3 of their own pieces. The line can be:

1. **Horizontal** (side-by-side in a row)
*Example of a horizontal win for Player 1 ('O'):*
```
X . .
O O O <-- 3 'O's in row 2
. X .
```
2. **Vertical** (stacked on top of each other in a column)
*Example of a vertical win for Player 2 ('X'):*
```
. X O
O X O <-- 3 'X's in column 2
. X .
```
3. **Diagonal** (connected at an angle)
*Example of a diagonal win (bottom-left to top-right) for Player 1:*
```
. . O
. O X <-- 3 'O's in a diagonal line
O X .
```
*Example of another diagonal win (top-left to bottom-right) for Player 2:*
```
X . O
. X O <-- 3 'X's in a diagonal line
. O X
```
**Draw Condition**:
If the entire grid is filled with pieces and no player has won, the game is a draw.

## Current Game State
{state_prompt}

## Your Turn
You are {player}.
The available actions are: {actions}.
\end{promptbox}

\subsection{Who's the Spy}

\textit{Who's the Spy} is a multi-player, multi-turn social deduction game involving $N$ players, among whom $N-1$ are assigned as \textit{Civilians} and one as the \textit{Undercover}. All players are secretly assigned a word: the civilians share the same word, while the undercover receives a different but semantically related word. Importantly, players do not know anyone's identity throughout the game.

The game proceeds in multiple speaking rounds. In each round, players take turns generating short descriptions of their assigned word, aiming to convey its meaning without revealing too much explicit information. After two full speaking rounds, all players simultaneously vote to eliminate one suspected undercover. The player receiving the most votes is removed from the game. The civilians win if the undercover is successfully eliminated, while the undercover wins if at least one civilian is voted out.

Winning \textit{Who's the Spy} requires a delicate balance between informativeness and concealment. Civilian players must describe the shared word accurately enough to signal alignment with other civilians, while avoiding overly explicit descriptions that could allow the undercover to infer the civilians' word and adapt accordingly. Conversely, the undercover must generate plausible but strategically ambiguous descriptions to blend in and avoid detection. This interaction demands players to reason about others' beliefs, intentions, and linguistic strategies, thereby strongly engaging theory-of-mind capabilities. At the same time, the open-ended nature of word description encourages flexible and creative language generation.

In our experiments, we fix the number of players to four, consisting of three civilians and one undercover, and set the Qwen3-14B model as training opponents. For each game instance, the identity and speaking order of the trained LLM are randomly assigned. A trajectory is considered successful if the side corresponding to the identity assigned to the trained model wins the game. The word list used in our experiments is adopted from an open-source resource\footnote{\url{https://github.com/xzx34/SocialMaze/tree/main/find_the_spy}}. To improve training robustness and reduce sensitivity to prompt phrasing, we design four distinct rule prompts that describe the game mechanics and player objectives from different perspectives. During training, one rule prompt is randomly sampled for each episode. An example of such a rule prompt is shown below:
\begin{promptbox}
Game: Who's the Undercover Agent
Roles:
- 3 Civilians share one word.
- 1 Undercover has a related but different word.  

Goal:
- Civilians: Find the undercover.
- Undercover: Stay hidden until only 2 players remain.

How to Play:
Each player describes their word in one sentence (without saying the word itself).
Be subtle yet clear. After two-turn speak, everyone votes out one player. The one with most votes is eliminated.
  
Win:
- Civilians win if the undercover is voted out.
- Undercover wins if one civilian is voted out.
\end{promptbox}

During the testing phase, we observe that the model tends to repeatedly generate highly similar or identical descriptions across turns, which reduces linguistic diversity and weakens the effectiveness of social deduction training. To mitigate this issue, we introduce an additional hint prompt during the description phase to encourage more varied and informative language generation: 
\begin{promptbox}
### Additional Rules for Description (Very Important)
- Your description MUST be clearly different from any descriptions you have given in earlier rounds.
- Do NOT reuse similar words, sentence structures, or ideas. Avoid describing it as a "process" again.
- Each round, pick a NEW angle of interpretation (e.g., its effect, its form, its symbolism, its usage, etc.)
- The new description should have LOW semantic similarity with your previous descriptions.
- Act as if you don't remember your previous answers, but you MUST ensure this answer is not similar to them.
- Always produce a new, creative, and distinct sentence.
- If the word has multiple meanings, assume 100\% that the basic meaning is intended. Never choose the less common or technical ones.
\end{promptbox}

\subsection{Derivation of CST Optimization Effect}
\label{app:seq-derivation}

We provide the detailed derivation of the second-order terms induced by CST. Let $\mathcal{L}^{(i)}_{\mathrm{RL}}(\theta)$ denote the local RL objective for subtask $i$, and define
\begin{equation}
    g_i(\theta)=\nabla_\theta \mathcal{L}^{(i)}_{\mathrm{RL}}(\theta),
    \qquad
    H_i(\theta)=\nabla_\theta^2 \mathcal{L}^{(i)}_{\mathrm{RL}}(\theta).
\end{equation}
For brevity, unless otherwise specified, we write $g_i=g_i(\theta)$ and $H_i=H_i(\theta)$, both evaluated at the initial parameter $\theta$ of the CST cycle.

Consider an arbitrary subtask order $\pi=(\pi_1,\ldots,\pi_K)$. A one-step CST cycle is
\begin{equation}
\begin{aligned}
    \theta_{\mathrm{CST}}
    &=
    U_{\pi_K}
    \circ
    U_{\pi_{K-1}}
    \circ
    \cdots
    \circ
    U_{\pi_1}
    (\theta),
    \\
    U_i(\theta)
    &=
    \theta - \eta g_i(\theta).
\end{aligned}
\end{equation}
Let $\theta_s$ be the parameter after the first $s$ local updates. Since
\begin{equation}
    \theta_{s-1}
    =
    \theta
    -
    \eta
    \sum_{r<s} g_{\pi_r}
    +
    O(\eta^2),
\end{equation}
Taylor expansion gives
\begin{equation}
\begin{aligned}
    g_{\pi_s}(\theta_{s-1})
    &=
    g_{\pi_s}(\theta)
    +
    H_{\pi_s}(\theta)(\theta_{s-1}-\theta)
    +
    O(\eta^2) \\
    &=
    g_{\pi_s}
    -
    \eta
    H_{\pi_s}
    \sum_{r<s} g_{\pi_r}
    +
    O(\eta^2).
\end{aligned}
\end{equation}
Substituting this into $\theta_s=\theta_{s-1}-\eta g_{\pi_s}(\theta_{s-1})$ and accumulating over $s=1,\ldots,K$ yields
\begin{equation}
    \theta_{\mathrm{CST}}
    =
    \theta
    -
    \eta
    \sum_{s=1}^{K} g_{\pi_s}
    +
    \eta^2
    \sum_{r<s}
    H_{\pi_s} g_{\pi_r}
    +
    O(\eta^3).
    \label{eq:app-cst-second-order-order}
\end{equation}
For the canonical order $(1,\ldots,K)$, this reduces to
\begin{equation}
    \theta_{\mathrm{CST}}
    =
    \theta
    -
    \eta
    \sum_{i=1}^{K} g_i
    +
    \eta^2
    \sum_{i<j} H_j g_i
    +
    O(\eta^3).
    \label{eq:app-cst-second-order}
\end{equation}
The first-order term is identical to the naive mixed update. The difference is the ordered second-order term $\sum_{i<j}H_j g_i$, which appears because later subtask gradients are evaluated at parameters already modified by earlier subtask updates.

We next relate this ordered term to gradient compatibility. Define the pairwise gradient-inner-product objective
\begin{equation}
    \mathcal{A}(\theta)
    =
    \sum_{i<j}
    g_i(\theta)^\top g_j(\theta).
\end{equation}
Assuming the local objectives are twice differentiable, the Hessians are symmetric, and therefore
\begin{equation}
\begin{aligned}
    \nabla_\theta \mathcal{A}(\theta)
    &=
    \sum_{i<j}
    \nabla_\theta
    \left(
    g_i(\theta)^\top g_j(\theta)
    \right) \\
    &=
    \sum_{i<j}
    \left(
    H_i g_j + H_j g_i
    \right).
\end{aligned}
\label{eq:app-inner-product-gradient}
\end{equation}
Thus, a symmetric combination of the two ordered interactions $H_j g_i$ and $H_i g_j$ corresponds exactly to the gradient of the pairwise inner product.

More generally, let $p_{ij}$ denote the probability, under an order distribution, that subtask $i$ appears before subtask $j$. Taking expectation over orders in Eq.~\ref{eq:app-cst-second-order-order}, the ordered interaction term becomes
\begin{equation}
\begin{aligned}
    \mathbb{E}_{\pi}
    \left[
    \sum_{r<s}
    H_{\pi_s} g_{\pi_r}
    \right]
    &=
    \sum_{i<j}
    \Big[
        p_{ij} H_j g_i
    \\
    &\qquad\qquad
        +
        (1-p_{ij}) H_i g_j
    \Big].
\end{aligned}
\label{eq:app-order-expected-term}
\end{equation}
When the order distribution is pairwise balanced, i.e., $p_{ij}=1/2$ for every pair $(i,j)$, we obtain
\begin{equation}
    \mathbb{E}_{\pi}
    \left[
    \sum_{r<s}
    H_{\pi_s}g_{\pi_r}
    \right]
    =
    \frac{1}{2}
    \nabla_\theta
    \mathcal{A}(\theta).
    \label{eq:app-balanced-alignment}
\end{equation}
This is the same second-order view used by Reptile~\cite{nichol2018reptile}: sequential gradient steps introduce higher-order terms that act like an update toward increasing gradient inner products. Since
\begin{equation}
    g_i^\top g_j
    =
    \|g_i\|
    \|g_j\|
    \cos(g_i,g_j),
\end{equation}
increasing the inner product encourages smaller angles between subtask gradients when their norms are comparable. This makes the corresponding policy-improvement directions more compatible.

Our implementation uses a fixed cyclic schedule, e.g., $A\!\rightarrow\!B\!\rightarrow\!C\!\rightarrow\!A\!\rightarrow\!B\!\rightarrow\!C$. For a single cycle with a fixed starting phase, Eq.~\ref{eq:app-cst-second-order} is the exact second-order expansion, and the interaction term is directional rather than fully symmetric. Therefore, Eq.~\ref{eq:app-balanced-alignment} should be understood as an order-averaged interpretation, not as an exact identity for one fixed cycle. In continuous cyclic training, however, reverse-direction interactions also occur across cycle boundaries. For a long cyclic sequence containing $N$ complete repetitions, each unordered pair contributes in both directions, and the imbalance between the two directions is only a boundary effect of order $O(N)$ among $O(N^2)$ cross-pair interactions. Thus, the repeated cyclic process provides a practical approximation to the pairwise-balanced view, while still preserving the exact ordered second-order interactions in each local cycle.

We finally extend the derivation to the block form used by CST. Let $U_i^{(\tau)}$ denote applying $U_i$ for $\tau$ consecutive local steps. For a single subtask block, Taylor expansion gives
\begin{equation}
    U_i^{(\tau)}(\theta)
    =
    \theta
    -
    \eta \tau g_i
    +
    \eta^2
    \frac{\tau(\tau-1)}{2}
    H_i g_i
    +
    O(\eta^3).
    \label{eq:app-block-single}
\end{equation}
Applying these blocks sequentially over $K$ subtasks yields
\begin{equation}
\begin{aligned}
    \theta_{\mathrm{CST}}
    =
    \theta
    &-
    \eta \tau
    \sum_{i=1}^{K} g_i \\
    &+
    \eta^2
    \frac{\tau(\tau-1)}{2}
    \sum_{i=1}^{K} H_i g_i \\
    &+
    \eta^2 \tau^2
    \sum_{i<j} H_j g_i
    +
    O(\eta^3).
\end{aligned}
\label{eq:app-cst-block-second-order}
\end{equation}
The additional within-subtask term reflects local consolidation inside each homogeneous block. The cross-subtask term is the same coordination effect derived above, scaled by $\tau^2$. When $\tau=1$, Eq.~\ref{eq:app-cst-block-second-order} reduces to the basic CST expansion in Eq.~\ref{eq:app-cst-second-order}.
\section{Detailed Training settings}
\label{app:experiment}

\subsection{Optimization Objective}

We adopt a trajectory-based reinforcement learning formulation using Group Relative Policy Optimization (GRPO) as the optimization backbone following the RAGEN framework\cite{wang2025ragen}. In multi-turn settings, the model interacts with the environment over a full trajectory $\tau_i$, and a scalar reward is assigned at the trajectory level. The resulting advantage is normalized and distributed across all token positions within the trajectory.

Given a group of $G$ sampled trajectories $\{\tau_i\}_{i=1}^{G}$, we define the policy ratio at token position $t$ as
\begin{equation}
r_{i,t}(\theta)
=
\frac{\pi_\theta(\tau_{i,(t)} \mid \tau_{i,<t})}
     {\pi_{\text{old}}(\tau_{i,(t)} \mid \tau_{i,<t})},
\end{equation}
and its clipped version as
\begin{equation}
\tilde{r}_{i,t}(\theta)
=
\text{clip}\!\left(
r_{i,t}(\theta),
1-\epsilon,\,1+\epsilon
\right).
\end{equation}

The trajectory-level GRPO objective is then given by
\begin{equation}
\begin{aligned}
\frac{1}{G}
\sum_{i=1}^{G}
\frac{1}{|\tau_i|}
\sum_{t=1}^{|\tau_i|}
\min \Big[
r_{i,t}(\theta)\hat{A}_{i,t},
\;
\tilde{r}_{i,t}(\theta)\hat{A}_{i,t}
\Big],
\end{aligned}
\end{equation}

where $\hat{A}_{i,t}$ denotes the normalized advantage at token position $t$ within trajectory $\tau_i$, and $\epsilon$ is the clipping threshold.

This objective follows a standard PPO-style formulation and serves as a unified optimization backbone for all training settings in this work. 

\subsection{Coordinated Subtask Training Framework}
\label{app:nested}

Coordinated Subtask Training (CST) reformulates multi-task RL by sequentially applying subtask-specific policy updates instead of optimizing all subtasks through a single mixed update. Specifically, a CST cycle consists of an ordered sequence of local updates, where each update uses trajectories sampled from one subtask only. Each subtask preserves its original interaction protocol, reward function, and success condition, while the model parameters are passed from one local update to the next.

This formulation differs from naive mixed training. In mixed training, trajectories from different subtasks are placed in the same batch and optimized through one aggregated policy update. This can contaminate task-local RL signals because advantages, entropy terms, and policy-gradient scales are calibrated over heterogeneous trajectory distributions. In contrast, CST estimates each local gradient within a homogeneous subtask-specific batch, yielding cleaner task-local optimization signals.

CST also changes how subtasks interact during optimization. Instead of directly summing gradients from different subtasks, CST applies them sequentially. Therefore, later subtask updates are computed on parameters already modified by earlier subtasks. This sequential structure introduces additional second-order cross-subtask terms, such as $\eta^2 H_j g_i$ for two subtasks and $\eta^2(H_B g_A + H_C g_A + H_C g_B)$ for three subtasks ordered as $A \rightarrow B \rightarrow C$. These terms are absent from direct gradient summation and provide an implicit mechanism for coordinating subtask optimization directions.

In practice, CST can be implemented with a block length $\tau$, where each subtask performs $\tau$ consecutive local updates before switching to the next subtask. The case $\tau=1$ recovers the basic CST cycle, while larger $\tau$ strengthens local consolidation within each homogeneous subtask block. Moreover, we use a fixed cyclic schedule over subtasks, e.g., $A \rightarrow B \rightarrow C \rightarrow A \rightarrow B \rightarrow C$, so that all subtasks are repeatedly optimized while maintaining clear subtask boundaries.

\subsection{Training Configuration and Implementation}

Following the RAGEN framework based on VeRL~\cite{sheng2024hybridflow}, we utilize the stable $\text{StarPO}^*$ algorithm. The key improvements in $\text{StarPO}^*$ are DAPO \cite{yu2025dapo} and trajectory filtering. DAPO removes the KL-term and employs a clip-higher strategy. The trajectory filtering is based on variance, where only the top 25\% of trajectories with the highest variance are retained for each round.

Regarding hyperparameters, each setting runs a maximum of 250 rollout-upodate iterations, with the group size fixed to 16. During training, we fixed the random seed to generate the task prompts to ensure reproducibility. We use $\tau=4$ for Qwen2.5-1.5B and $\tau=1$ for Qwen2.5-7B as fixed training settings across all CST experiments for each model scale. Exploring adaptive or task-dependent block lengths is left for future work.

The maximum number of turns for each sub-task is determined by the nature of the task. For single-turn tasks such as \textit{Math} and \textit{Matrix Games}, the maximum number of turns is 1. For multi-turn tasks, the maximum turns for \textit{TicTacToe} is set to 5, since the game involves a 3x3 grid, and the game will always end after the 5th turn. In \textit{Who's the Spy}, the maximum turns is 3, consisting of two description rounds and one voting round. In mixed training and CST, the \texttt{max\_turn} is set to the maximum number of turns across all sub-tasks. 

The model is optimized using Generalized Advantage Estimation (GAE) with $\gamma = 1.0$ and $\lambda = 1.0$, along with the Adam optimizer where $\beta_1 = 0.9$ and $\beta_2 = 0.999$. We also apply entropy regularization with a coefficient $\beta = 0.001$. The reward is defined as 1 for success and 0 for failure, with a reward of 0.5 assigned to draws in \textit{TicTacToe}. The format penalty of $-0.1$ is applied. 

In multi-task settings, both mixed training and CST scenarios involve an equal number of tasks, ensuring fairness in task comparison.

\section{Detailed Evaluation Settings}
\label{app:common}
We introduce the detailed evaluation settings including math, game and general abilities scenarios.
\subsection{Math Evaluation}
We evaluate the model's math reasoning ability using the MATH500 benchmark. Model outputs are compared against ground-truth answers using the Math-Verify toolkit\footnote{\url{https://github.com/huggingface/Math-Verify}}, which provides a robust verification pipeline for mathematical equivalence and correctness.

\subsection{Game Evaluation}
To assess game-playing ability, we evaluate the model's success rate over 100 independent game rounds, with the opponent model fixed to Gemini-2.5-Flash for \textit{TicTacToe} and \textit{Who's the Spy} tasks. For each game, the success criterion is defined in accordance with the corresponding training objective. Specifically, in \textit{Matrix Games}, a trajectory is considered successful if the selected action satisfies the Nash Equilibrium condition. In \textit{TicTacToe}, success is defined as achieving a valid three-in-a-row configuration (horizontal, vertical, or diagonal) for the trained model's pieces. To make the evaluation of \textit{TicTacToe} fair, models are tested on an empty board and the first player is randomly selected. In \textit{Who's the Spy}, a trajectory is counted as successful if the side corresponding to the trained model's assigned identity wins the game. 

\subsection{General Abilities Evaluation}
To evaluate the model's generalization across diverse ability dimensions, we select a set of widely used benchmarks, including MMLU, MMLU-Pro, CommonGen, and SocialIQA.

\paragraph{MMLU and MMLU-Pro.}
MMLU covers a broad range of academic subjects and evaluates knowledge and reasoning through four-choice questions. MMLU-Pro extends MMLU by increasing both task difficulty and answer space, expanding each question from four options to ten options, thereby posing a more challenging evaluation setting.

\paragraph{CommonGen.}
The CommonGen benchmark is designed to evaluate the model's language generation ability by requiring it to produce a coherent and plausible sentence that incorporates a given set of concept words with specified semantic roles, as illustrated in Fig.~\ref{fig:case-mnc-common}. Since CommonGen is an open-ended generation task, we follow the evaluation protocol provided by allenai\footnote{\url{https://github.com/allenai/CommonGen-Eval}}, using few-shot prompting and GPT-4o as an automatic evaluator to compare model outputs against human-annotated reference sentences. To reduce positional bias during judgment, we randomly permute the order of the model-generated output and the reference sentence, and instruct the evaluator to carefully compare both candidates. Since the human-annotated references represent a strong upper bound that is difficult to consistently surpass, we report the success rate using a win-or-tie criterion. The detailed evaluation prompt is provided below:
\begin{promptbox}
# Data
Given several concepts (i.e., nouns or verbs), we ask models to write a short and simple sentence that contains *all* the required words.
The sentence should describe a common scene in daily life, and the concepts should be used in a natural way.
Concepts: "{concept_list}"
Model A: "{candidate_A}"
Model B: "{candidate_B}"

# Your Task
Your task is to choose a better sentence from the two candidates. Decide which model's sentence is better in terms of the naturalness and commonness of the scenes they describe.

## Rules:
- A better sentence should describe a common scene in daily life, and all concepts should be used in a natural way.
- You should prefer sentences that use all given concepts with correct part-of-speech tags.
- A simpler and shorter sentence is preferred if it describes the same scene as the other sentence.
- If you think both sentences are equally good or bad, please choose "tie".

Now, please output your choice ("A" or "B" or "tie").

Your choice:
\end{promptbox}

\paragraph{SocialIQA.}
SocialIQA is a question-answering benchmark designed to evaluate social reasoning through three-choice questions. Unlike prior benchmarks that emphasize physical or taxonomic knowledge, SocialIQA focuses on understanding people's actions and their social motivations. Given a described action, the model must infer the most plausible social intent or implication among multiple candidates. The dataset covers a wide range of everyday social situations, with answer options consisting of both human-written and adversarially filtered machine-generated candidates. 

\begin{table*}[t]
\centering
\small
\setlength{\tabcolsep}{3pt}
\renewcommand{\arraystretch}{1.0}
\begin{tabular}{@{}l l c c c c | c c c c c@{}}
% \hline
% \multirow{2}{*}{Setting}
% & \multirow{2}{*}{Model}
% & \multicolumn{4}{c}{\textbf{In-Domain Tasks}}
% & \multicolumn{4}{c}{\textbf{Out-of-Distribution Tasks}} \\
% \cmidrule(lr){3-6} \cmidrule(lr){7-10}
% & 
\toprule
\makecell[c]{\textbf{Setting}}
& \makecell[c]{\textbf{Opponent}}
& \makecell[c]{\textbf{MATH}}
& \makecell[c]{\textbf{Matrix}}
& \makecell[c]{\textbf{TicTacToe}}
& \makecell[c]{\textbf{Spy}}
& \makecell[c]{\textbf{MMLU}}
& \makecell[c]{\textbf{MMLU-Pro}}
& \makecell[c]{\textbf{Common}}
& \makecell[c]{\textbf{Social}} 
& \makecell[c]{\textbf{Avg.}}\\
\midrule

\multirow{2}{*}{\textit{TicTacToe}}
& Gemini-2.5-Flash
& 25.80 & 17.00 & \cellcolor{grey_purple} 46.00 & 12.00 & 39.00 & 13.79 & 14.32 & 51.18 & 29.57 \\
& \textbf{Qwen3-14B}
& 22.40 & 32.00 & \cellcolor{grey_purple} 75.00 & 21.00 & 47.89 & 16.45 & 11.56 & 64.12 & 35.00 \\
\midrule

\multirow{2}{*}{\textit{Who's the Spy}}
& Gemini-2.5-Flash
& 19.20 & 19.00 & 2.00 & \cellcolor{grey_purple} 38.00 & 46.86 & 17.21 & 20.85 & 59.57 & 36.12 \\
& \textbf{Qwen3-14B}
& 19.20 & 21.00 & 0.00 & \cellcolor{grey_purple} 33.00 & 43.85 & 15.68 & 20.85 & 60.70 & 35.27 \\

\bottomrule
\end{tabular}
\caption{Opponent sensitivity analysis of \textit{TicTacToe} and \textit{Who's the Spy} training for base 1.5B model, comparing Gemini-2.5-Flash and Qwen3-14B as opponents.}
\label{tab:ablation_opponent}
\end{table*}

% \section{Detailed Ablation Study}
% The detailed ablation study tables for opponent sensitivity analysis is shwon iTable~\ref{table:ablation_opponent}.

\section{Detailed Case Studies}
\label{app:case-study}

\paragraph{Effectiveness of Informal Learning on Mathematical Reasoning.}

\begin{figure*}[t]
  \centering
  \includegraphics[width=\textwidth]{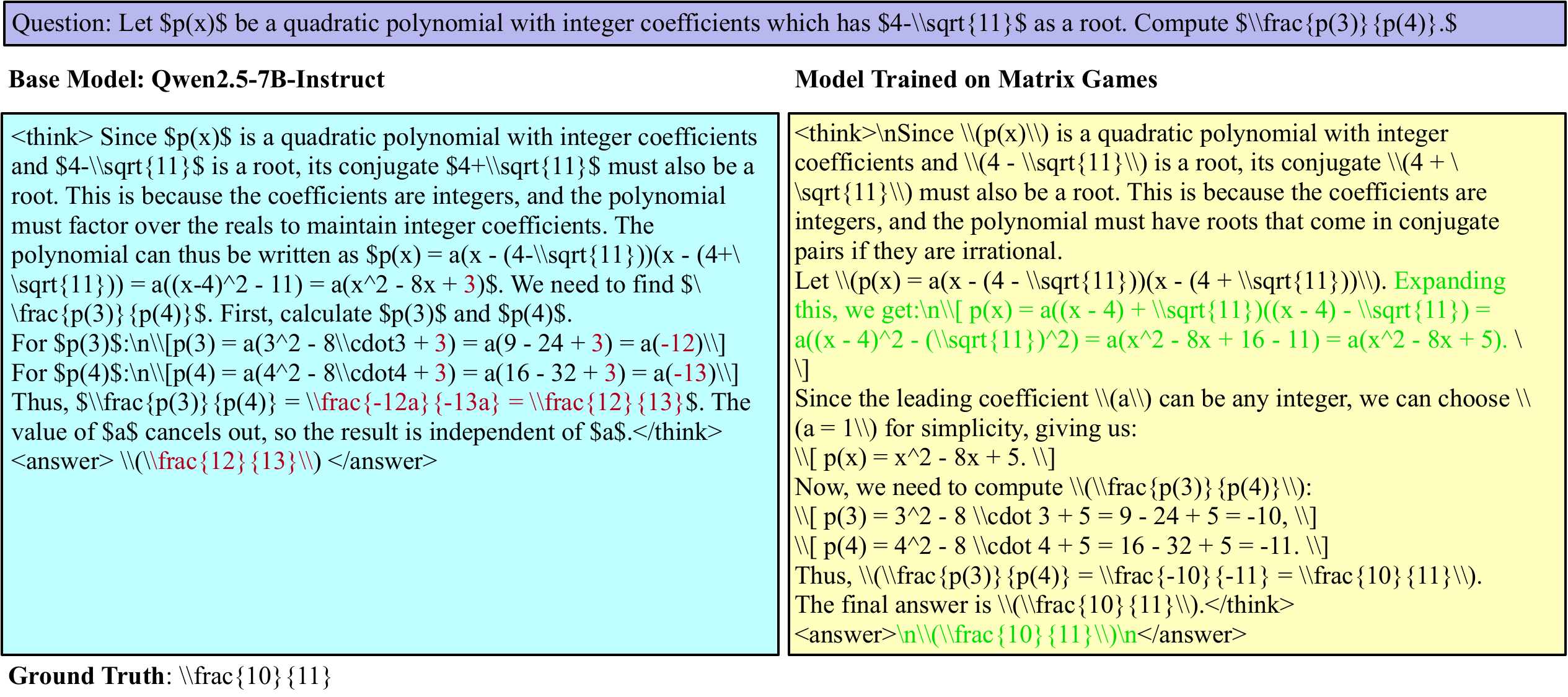}
  \caption{Detailed case study on MATH500, comparing 7B base model trained on Math with \textit{Matrix Games}.}
  \label{fig:case-matrix-math}
\end{figure*}

To provide a more detailed illustration of how informal learning improves mathematical reasoning, we present a representative case study on the MATH500 benchmark in Fig.~\ref{fig:case-matrix-math}. The example compares the base \textit{Qwen2.5-7B-Instruct} model with the same model after training on \textit{Matrix Games}.

In this example, both models correctly identify that if a quadratic polynomial with integer coefficients has $4-\sqrt{11}$ as a root, then its conjugate $4+\sqrt{11}$ must also be a root. The base model further recognizes the high-level structure of the solution and attempts to construct the corresponding polynomial. However, it commits a subtle arithmetic error when expanding $(x-4)^2 - 11$, incorrectly computing $16 - 11$ as $3$. This local mistake propagates to the subsequent evaluation of $p(3)$ and $p(4)$, ultimately leading to an incorrect final answer, despite the overall reasoning strategy being correct.

In contrast, the model trained with \textit{Matrix Games} exhibits a more structured and explicit derivation. Instead of directly simplifying intermediate expressions, it expands the polynomial step by step as
\[
(x-4)^2 - 11 = x^2 - 8x + 16 - 11,
\]
making the critical arithmetic operation transparent and easy to verify. Moreover, it explicitly observes that the leading coefficient $a$ can be any integer and chooses $a=1$ for simplicity, demonstrating greater flexibility and creativity in solution construction. This explicit and verifiable reasoning process helps the model avoid local arithmetic errors and arrive at the correct final result. This case suggests that informal learning through interactive game environments encourages more cautious, structured, and self-checking reasoning behaviors in mathematical problem solving.

\paragraph{Effectiveness of CST on Semantic Generation.}

\begin{figure*}[t]
  \centering
  \includegraphics[width=\textwidth]{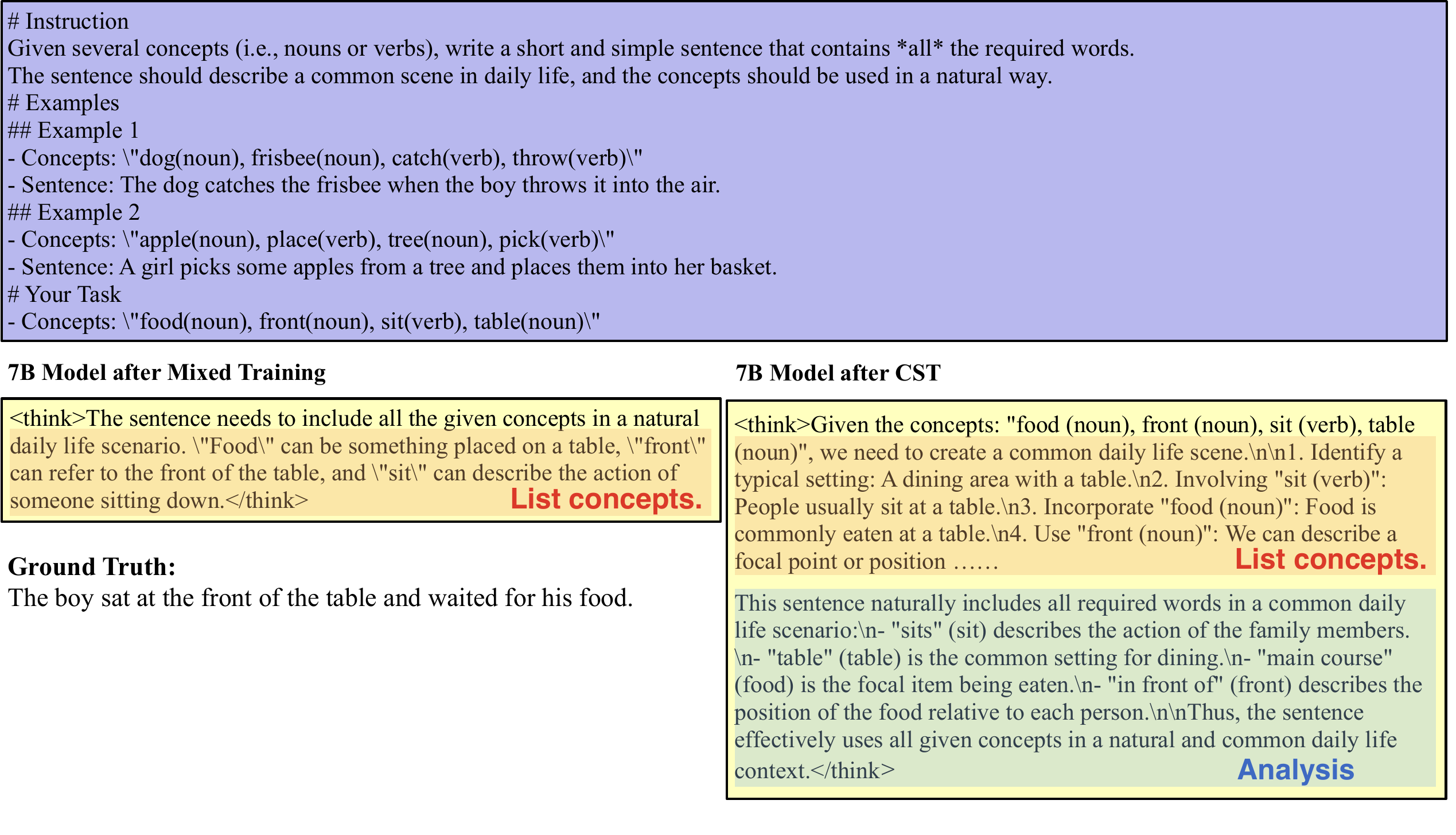}
  \caption{Detailed case study on CommonGen, comparing 7B base model on F+$\text{I}_2$ settings mixed training with CST.}
  \label{fig:case-mnc-common}
\end{figure*}

We further analyze the difference between mixed and CST strategies through a detailed case study on the CommonGen benchmark, as shown in Fig.~\ref{fig:case-mnc-common}. The task requires generating a coherent daily-life sentence that naturally incorporates all given concepts.

The model trained with mixed objectives is able to correctly include the required concepts and produce a grammatically valid sentence. However, its generation primarily focuses on satisfying individual lexical constraints, resulting in a relatively minimal and less vivid description. While the output is semantically acceptable, it lacks explicit scene construction and deeper integration among the concepts.

In contrast, the model trained with CST demonstrates a more deliberate generation process. Before producing the final sentence, it briefly verifies the required concepts and analysis its answer. This leads to a more expressive, contextually rich, and human-aligned sentence. As a result, the CST model achieves a win-or-tie outcome when compared against strong human-annotated references.

This comparison highlights that CST encourages deeper semantic integration and scene-level reasoning, going beyond word-level constraint satisfaction. By requiring the model to jointly consider multiple objectives within a single trajectory, CST promotes more coherent and expressive language generation than naive mixed training.

\begin{table*}[t]
\centering
\small
\setlength{\tabcolsep}{3pt}
\renewcommand{\arraystretch}{1.0}
\begin{tabular}{@{} l c c c c | c c c c c@{}}
% \hline
% \multirow{2}{*}{Setting}
% & \multirow{2}{*}{Model}
% & \multicolumn{4}{c}{\textbf{In-Domain Tasks}}
% & \multicolumn{4}{c}{\textbf{Out-of-Distribution Tasks}} \\
% \cmidrule(lr){3-6} \cmidrule(lr){7-10}
% & 
\toprule
\makecell[c]{\textbf{Model}}
& \makecell[c]{\textbf{MATH}}
& \makecell[c]{\textbf{Matrix}}
& \makecell[c]{\textbf{TicTacToe}}
& \makecell[c]{\textbf{Spy}}
& \makecell[c]{\textbf{MMLU}}
& \makecell[c]{\textbf{MMLU-Pro}}
& \makecell[c]{\textbf{Common}}
& \makecell[c]{\textbf{Social}} 
& \makecell[c]{\textbf{Avg.}}\\
\midrule
GPT-4o
& 59.80 & 48.00 & 63.00 & 70.00 & 85.08 & 66.00 & 53.02 & 79.02 & 70.78 \\
Gemini3-Flash
& 49.80 & 64.00 & 86.00 & 16.00 & 78.42 & 52.01 & 41.96 & 74.87 & 61.82 \\
DeepSeek-V3.2 & 55.80 & 61.00 & 88.00 & 69.00 & 86.73 & 63.69 & 67.34 & 80.19 & 74.49  \\

\bottomrule
\end{tabular}
\caption{Additional results of close-source models. }
\label{tab:add_api}
\end{table*}

\section{Detailed Ablation Studies}
\label{app:opp}
\subsection{Opponent Sensitivity in Multi-player Games}
We study the effect of opponent choice by training the 1.5B model against different opponents, as shown in Table~\ref{tab:ablation_opponent}. Overall, the choice of opponent mainly affects the game on which the model is trained, while its impact on general benchmark performance is relatively small.

For \textit{TicTacToe} training, using Qwen3-14B as the opponent substantially improves the \textit{TicTacToe} success rate from 46\% to 75\%, yielding a 29-point absolute gain over Gemini-2.5-Flash. This suggests that training against the stronger Qwen3-14B opponent encourages the model to learn more robust game-specific strategies. In addition, the average score increases from 29.57 to 35.00, indicating that the stronger opponent does not degrade overall capability. Although performance on some non-game benchmarks slightly decreases, such as MATH from 25.80 to 22.40 and Common from 14.32 to 11.56, the model obtains higher scores on Matrix, MMLU, MMLU-Pro, and Social, showing no clear evidence of broad negative transfer.

For \textit{Who's the Spy} training, the difference between opponents is smaller. Training against Gemini-2.5-Flash achieves a higher \textit{Who's the Spy} success rate of 38\%, compared with 33\% when training against Qwen3-14B. However, Qwen3-14B still provides competitive overall performance, with an average score of 35.27 compared with 36.12 for Gemini-2.5-Flash. The two settings also show similar performance on most general benchmarks, suggesting that opponent choice has a weaker effect in \textit{Who's the Spy} than in \textit{TicTacToe}. 

Notably, Qwen3-14B outperforms Gemini-2.5-Flash in direct evaluations, achieving success rates of 66\% in \textit{TicTacToe} and 52\% in \textit{Who's the Spy}. Together with its strong training performance in \textit{TicTacToe} and the practical advantage that the open-sourced Qwen3-14B can be deployed locally with lower inference latency, we adopt Qwen3-14B as the default opponent in all experiments.

\section{Additional Results of Closed-Source models}
\label{app:api}

% We further conduct experiments on strong closed-source models, including GPT-4o, Gemini3-Flash-Preview and DeepSeekV3.2(chat mode, without thinking) \cite{liu2025deepseek}. The evaluation methods are aligned with the main experiments. Results are shown in Table.~\ref{tab:add_api}.

% We further evaluate a set of strong closed-source models, including GPT-4o, Gemini-3-Flash-Preview, and DeepSeek-V3.2 (chat mode, without reasoning)~\cite{liu2025deepseek}. All evaluation protocols are strictly aligned with those used in the main experiments. The corresponding results are reported in Table~\ref{tab:add_api}.
We further evaluate a set of strong closed-source models, including GPT-4o, Gemini-3-Flash-Preview, and DeepSeek-V3.2 (chat mode, without reasoning)~\cite{liu2025deepseek}. All evaluation protocols are strictly aligned with those used in the main experiments. The corresponding results are reported in Table~\ref{tab:add_api}.

Overall, these results demonstrate that strong closed-source models exhibit competitive performance across both formal and informal learning benchmarks, while showing distinct strengths on different task categories. GPT-4o achieves balanced performance across mathematical reasoning, social reasoning, and general knowledge benchmarks, reflecting its strong general-purpose capability. Gemini-3-Flash-Preview performs particularly well on interactive game tasks such as \textit{TicTacToe} and \textit{Matrix Games}, but shows relatively weaker performance on social deduction and general language generation tasks. DeepSeek-V3.2 achieves the highest overall average score, with consistently strong results on math, game, and general ability benchmarks, indicating that competitive generalization can be achieved even without explicit reasoning traces.These observations suggest that while closed-source models benefit from large-scale training and strong base capabilities, their performance profiles vary substantially across different ability dimensions. 

Moreover, we observe that the 1.5B and 7B models after RL training can, under certain domain-specific or CST settings, achieve performance comparable to or even surpass strong closed-source models on targeted benchmarks. As shown in the main results, a 1.5B model trained under the $\text{F + I}_1$ setting attains 65.00\% accuracy on \textit{Matrix Games}, exceeding the performance of all evaluated closed-source models in this domain. 

Similar phenomena are also observed for 7B models on several general ability benchmarks, including MMLU-Pro, CommonGen, and SocialIQA, where models trained with CST or domain-aligned tasks achieve results on par with those of closed-source APIs. These gains are particularly pronounced on tasks that require structured reasoning, interaction, or social inference, which are directly targeted by the corresponding informal learning environments.

Taken together, these results indicate that, despite the substantially smaller model size and more limited training data, our proposed training framework can effectively elicit strong domain-specific and transferable capabilities through structured and CST reinforcement learning. This comparison with closed-source models provides further evidence that carefully designed interactive training objectives can partially compensate for scale and serve as an efficient alternative for capability enhancement.

\end{document}